\documentclass[10pt,twocolumn,letterpaper]{article}

\usepackage{cvpr}
\usepackage{times}
\usepackage{epsfig}
\usepackage{graphicx}
\usepackage{amsmath}
\usepackage{amssymb}
\usepackage{mathtools}
\usepackage{mathrsfs} % new
\usepackage{soul}
\usepackage{booktabs}
\usepackage{xcolor,pifont}
\newcommand*\colourcheck[1]{%
  \expandafter\newcommand\csname #1check\endcsname{\textcolor{#1}{\ding{52}}}%
}
\colourcheck{blue}
\colourcheck{green}
\colourcheck{red}
\newcommand{\xmark}{\ding{55}}%
\usepackage{titling}
\usepackage{multirow} 
\usepackage{enumitem}
\usepackage{color}

\newcommand{\doublecheck}[1]{\textcolor{black}{#1}}
\newcommand{\keypoint}[1]{\vspace{0.1cm}\noindent\textbf{#1}\quad}
\newcommand{\cut}[1]{}
\usepackage{algorithm}
\usepackage{algpseudocode}

\usepackage[nopar]{lipsum}
\usepackage{arydshln}
\makeatother
\usepackage[pagebackref=true,breaklinks=true,colorlinks,bookmarks=false]{hyperref}

\cvprfinalcopy % *** Uncomment this line for the final submission

\def\cvprPaperID{2112} % *** Enter the CVPR Paper ID here

\ifcvprfinal\fi
\begin{document}

%%%%%%%%% TITLE
\title{Doodle It Yourself: Class Incremental Learning by Drawing a Few Sketches}
% Authors at the same institution
% Authors at the same institution
\author{
Ayan Kumar Bhunia\textsuperscript{1} \hspace{.3cm}  
Viswanatha Reddy Gajjala\thanks{Interned with SketchX}  \hspace{.3cm}
Subhadeep Koley\textsuperscript{1,2}  \hspace{.2cm} 
Rohit Kundu\footnotemark[1]  \hspace{.2cm} \\
Aneeshan Sain\textsuperscript{1,2}  \hspace{.2cm}   
Tao Xiang\textsuperscript{1,2}\hspace{.2cm}  
Yi-Zhe Song\textsuperscript{1,2} \\
\textsuperscript{1}SketchX, CVSSP, University of Surrey, United Kingdom.  \\
\textsuperscript{2}iFlyTek-Surrey Joint Research Centre on Artificial Intelligence.\\
{\tt\small \{a.bhunia, s.koley, a.sain, t.xiang, y.song\}@surrey.ac.uk; viswanathareddy998@gmail.com} 
\vspace{-1.0cm}
}
\date{}

\maketitle
\ifcvprfinal\thispagestyle{empty}\fi

%%%%%%%%% ABSTRACT
\begin{abstract}
\vspace{-0.1cm}
The human visual system is remarkable in learning new visual concepts from just a few examples. This is precisely the goal behind few-shot class incremental learning (FSCIL), where the emphasis is additionally placed on ensuring the model does not suffer from ``forgetting''. {In this paper, we push the boundary further for FSCIL by addressing} two key questions that bottleneck its ubiquitous application (i) can the model learn from diverse modalities other than just photo (as humans do), and (ii) what if photos are not readily accessible (due to ethical and privacy constraints). Our key innovation lies in advocating the use of sketches as a new modality for class support. The product is a ``Doodle It Yourself" (DIY) FSCIL framework where the users can freely sketch a few examples of a novel class for the model to learn to recognise photos of that class. For that, we present a framework that infuses (i) gradient consensus for domain invariant learning, (ii) knowledge distillation for preserving old class information, and (iii) graph attention networks for message passing between old and novel classes. We experimentally show that sketches are better class support than text in the context of FSCIL, echoing findings elsewhere in the sketching literature.

% on this resemblance to human vision and put forward a new FSCIL variant that lets the model learn from other modalities other than just photo.  removes practical limitations 

% Machine learning systems often fail to mimic the human visual system ability to learn from fewer annotated data. This gap fuels interest in few-shot class-incremental learning (FSCIL). FSCIL aims to design models than can incrementally learn novel class representations with a few samples, without forgetting the the previously learnt ones. FSCIL requires a few photo samples from the novel classes during the incremental learning stage. Using user-given photos for novel-class representation learning may raise privacy or intellectual property violation issues. Therefore, building an FSCIL model without violating the data privacy norm is instrumental. In this paper, we aim at incrementally learning to classify unseen class photos without accessing user's personal photo samples. To attain this, we propose a novel usage of user-drawn sketches as a substitute to the conventional photos to incrementally learn and adapt to the test domain (photo). We present a framework that links Knowledge Distillation, Gradient Consensus, and Graph Attention Networks to handle this practical problem. The proposed framework mitigates catastrophic forgetting, and learns domain invariant features, while simultaneously generating more refined boundaries for the new classes. Comprehensive results demonstrate that the proposed method transcends baselines by a large margin on the Sketchy dataset.

\end{abstract}

\vspace{-0.6cm}
%%%%%%%%% BODY TEXT
\section{Introduction}
\vspace{-0.1cm}

Fully supervised learning has served us great with performances on ImageNet already surpassing human-level \cite{he2015delving}. In reality, however, such progress is primarily limited to a small number of object classes where labels were explicitly curated (1000 in ImageNet vs. possibly millions out there). Class Incremental Learning \cite{li2017learning, hsu2018re, kirkpatrick2017overcoming} is one of the popular fronts that attempt to extend model perception to novel classes while not ``forgetting" about classes learned already. Amongst its many variants, the \cut{very} recent Few-Shot Class Incremental Learning (FSCIL) \cite{tao2020few} is the most realistic where it also dictates the model to learn new classes with {{very few}} examples, the same as humans do.

\begin{figure}[t]
\label{fig1}
% \halign{-02cm}
\includegraphics[width=1\linewidth]{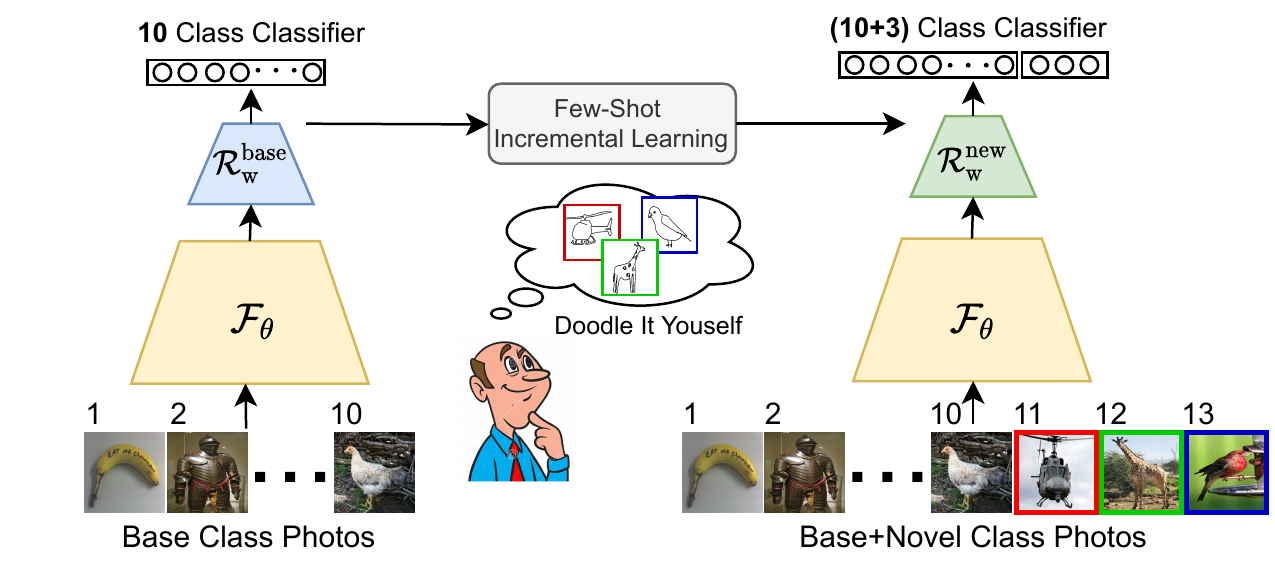}
\vspace{-0.5cm}
\caption{Illustration of our DIY-FSCIL framework. For instance, given \emph{sketch} exemplars (1-shot here) from $3$ novel classes as support-set, a $10$-class classifier gets updated to $(10+3)$-class classifier that can classify photos from both base and novel classes.}
% I think its better to use different images in the query instead of same images(same class diff images)

% \caption{Illustration of the proposed DIY framework for SB-FSCIL. The proposed system learns new classes progressively with the support from a few imaginative sketches doodled by users. Similar to us humans, our model learns from different modalities and addresses the privacy concerns by offering a substitute to the conventional data modality (photos).  }
\vspace{-0.5cm}
\end{figure}

% % Particulars of the problem we are solving - SUBHO 26-10-2021
As easy as providing a few samples might sound, questions start to emerge in practice as to (i) what data modality should the samples take? and (ii) how could these samples be obtained in practice. These questions, we argue, are key to the potentially ubiquitous application of FSCIL as (i) humans also learn from a broad range of data modalities that are not limited to just photo, and (ii) there are scenarios where photos are not necessarily always readily available due to privacy and ethical constraints (e.g., copyright). 

In this paper, we set out to study the role of human sketches as a support modality for FSCIL. This results in a flexible FSCIL system that learns new classes just by observing a few sketches \emph{doodled} by users themselves. Fig.~\ref{fig1} illustrates schematically our ``Doodle It Yourself (DIY)" FSCIL scenario -- ``DIY-FSCIL". {This importantly addresses the aforementioned problems in that (i) learning is no longer fixed to just photos but flexibly cross-modal with other data forms (just as humans do),} and (ii) it works without asking the users to source photos which might have practical constraints attached (e.g., copyright, hazardous environments). There is of course also the added benefit of injecting creativity to the classifier by sketching something off the user's imagination \cite{ge2020creative}, e.g., a ``flying cow"?

The advocate of sketches is largely motivated by the \cut{great} line of work examining human-centric characteristics of sketches in many parallel applications -- notably image retrieval \cite{dey2019doodle}, where the fine-grained nature of sketches \cut{(faithful yet flexible representation of a matching photo)} is used to successfully conduct instance-level retrieval \cite{bhunia2021more, bhunia2020sketch, sain2021stylemeup, dutta2019semantically, collomosse2019livesketch}.
Sketches in context of FSCIL is closely reminiscent of its usage in fine-grained retrieval. While in retrieval they utilise the detailed nature of sketches to conduct sketch-photo matching, we use a few sketches collectively as faithful visual representatives (support) of novel classes for incremental learning.
% Sketches in context of FSCIL closely follow its usage in fine-grained retrieval -- leveraging visually faithful nature of sketches to conduct sketch based incremental classification, by using a few sketches collectively as visual representatives (support) of novel classes.
% Modified OUrs
% Comparing sketches with text as class support, we find the former to be superior -- again thanks to its inherent fine-grained nature, echoing findings elsewhere in sketch-based image retrieval \cite{song2017fine}.
% modified ours
We show that sketches are better suited as class support in comparison to text, thanks to its inherent fine-grained nature,  validated by findings in contemporary sketch literature \cite{song2017fine, collomosse2019livesketch, bhunia2020sketch}.

%  echoing findings elsewhere in sketch-based image retrieval

% % Why is it non-trivial/ill-posed? - SUBHO

% #@YZS
Nonetheless, using sketches as class support in the FSCIL setting is non-trivial. \emph{Sketch}, despite being visually representative, is just a coarse contour-like depiction of the visual world, that sit in an entirely different domain from photo \cite{li2019episodic}. Thus, off-the-shelf models naively pre-trained on photos commonly fail to generalise well on sketches \cite{chen2019closer}. Moreover, due to its highly abstract nature, the same object may be sketched in various ways under unique user-styles \cite{song2018learning, sain2021stylemeup}, and with varied levels of detail \cite{sain2020cross}. We are also distinctly different to the parallel problem of SBIR -- SBIR typically get exposed to \textit{paired sketch-photo} data at training to learn a cross-modal embedding; we on the other hand need to work with \textit{sketches only} at training (i.e., no photo information whatsoever), yet still aim to generate classification layer weights to classify \textit{photos} from novel classes. 
% All these makes it crucial for our sketch-based FSCIL to learn a domain-invariant representation, workable across modalities.

% AB
% Three key design considerations present themselves for
Three key design considerations for  this
% this problem of 
cross-domain sketch-based FSCIL are: (i) how to make the model work cross-modal, (ii) how to preserve old class information, and (iii) how to leverage information from old classes to learn new ones. For the first issue, we design a gradient consensus based strategy that updates the model towards mutual agreement in the gradient space between sketch and photo domain, thus achieving a domain invariant feature extractor.  
For the second, we model an additional knowledge distillation loss to retain the acquired knowledge from old classes while incrementing the classifier to novel classes. Lastly, we devise a graph neural network to generate more discriminative decision boundaries for the incremented classifier via message passing between old and novel classes.

To summarise, our contributions are: (a) We extend \cut{the general line of} incremental learning research even further towards practicality and human-likeness. (b) {We achieve that by introducing sketches as class support for FSCIL, allowing the system to learn from modalities other than just photos and addressing issues around ethics and privacy while allowing user creativity.} (c) We introduce the first cross-modal framework to tackle this novel DIY-FSCIL problem.

% A gradient consensus-based approach on domain-invariant data. (b) A Graph Neural Network is embedded in the meta-learning stage for learning better decision boundaries on the novel class data. (c) Along with the traditional cross-entropy loss for model training, a knowledge distillation approach has been employed to mitigate the catastrophic forgetting problem of traditional CIL methods.

% Consistency regularisation ensures that the predictions generated by the learned classifier weights remain unaffected --> compared to the predictions made by the weight vectors generated from the  normal training.

% We aim to help novice users quickly add new classes to their models without domain knowledge and a tedious data collection process. We achieve it using our proposed cross-domain few-shot learning method and a sketching interface.

% Two constraints make our problem statement particularly challenging.

% Its not realistic to expect thousands of sketches from a user; instead we aim to incrementally learn a model using as few sketches as possible. 

% Its not possible to supply photos by the users due to privacy concerns(especially in the field of bio-metrics). In this work, we aim to provide a solution to these privacy concerns using sketches. 

\vspace{-0.3cm}
\section{Related Work}
\vspace{-0.1cm}
% Order of keypoints:
% ===================
% Sketch for Vision Tasks
% Sketch Based Image Retrieval
% Few-Shot Learning
% Incremental Learning
% Few-Shot Class-Incremental Learning
% Minimising Domain Discrepancy
\noindent \textbf{Sketch Based Image Retrieval (SBIR):} SBIR aims at retrieving paired photo given a query sketch, either at a \emph{category-level} \cite{collomosse2019livesketch, yelamarthi2018zero, dey2019doodle, ribeiro2020sketchformer, xu2018sketchmate} or at a finer-grained \emph{instance level} (FG-SBIR) \cite{sain2021stylemeup, PartialSBIR, strokesubset, bhunia2020sketch, bhunia2021more}. For learning the joint embedding space, category-level SBIR typically employs either CNN \cite{collomosse2019livesketch, dey2019doodle}, RNN \cite{xu2018sketchmate}, or Transformer \cite{ribeiro2020sketchformer} based Siamese networks, accompanied by a triplet-ranking objective \cite{yu2016sketch}. Contemporary research on this category is also directed towards zero-shot SBIR \cite{dey2019doodle,yelamarthi2018zero, Sketch3T} and binary hash-code embedding \cite{liu2017deep,shen2018zero}. \cut{ for learning generalizability and ease of computation respectively.} On the other hand, in FG-SBIR category, the seminal work by Yu \etal \cite{yu2016sketch} first introduced deep triplet-ranking based Siamese networks for joint embedding space learning, which was further reinforced by attention \cite{song2017deep}, cross-domain translation \cite{pang2017cross},\cut{textual query tags \cite{song2017fine},} reinforcement learning based on-the-fly retrieval \cite{bhunia2020sketch}, semi-supervised retrieval \cite{bhunia2021more}, style-agnostic retrieval \cite{sain2021stylemeup}, etc.

% In a completely different direction, we utilize sketches to address privacy concerns, learn new novel classes in a few shot incremental fashion, and reduce the time-consuming annotation process.

\noindent \textbf{Sketch for Vision Tasks:} Hand-drawn sketches, by nature, are enriched with various human visual system-like understanding abilities and are quite close to the cognitive-subconscious of human intelligence \cite{hertzmann2020line}. Consequently, it has facilitated various visual understanding tasks in the past. Apart from the widely studied SBIR \cite{collomosse2019livesketch, yelamarthi2018zero, dey2019doodle, ribeiro2020sketchformer, sain2021stylemeup, pang2019generalising, song2017deep, bhunia2020sketch, bhunia2021more}, sketch has also been employed in a variety of vision understanding tasks, including segmentation \cite{hu2020sketch}, video synthesis \cite{li2021deep}, representation learning \cite{wang2021sketchembednet, bhunia2021vectorization}, object localisation \cite{tripathi2020sketch}, image-inpainting \cite{xie2021exploiting}, 3D shape retrieval \cite{luo2020towards}, 3D shape modelling \cite{zhang2021sketch2model}, among others \cite{xu2022deep}. Some artistic application of sketch includes image editing \cite{yang2020deep}, animation auto-completion \cite{xing2015autocomplete}, etc. Sketches have lately been used to create Pictionary-style competitive drawing games \cite{sketchxpixelor}. These establish the fact that hand-drawn sketches have enough representative ability to characterise a visual photo efficiently. Set upon this fact, in this paper, we aim to explore how \emph{sketch} can act as a potential substitute to the conventional photos in class incremental learning.

% Chen et al. \cite{chen2018sketchygan} developed the ``SketchyGAN" framework that synthesizes the real world photos that match poses and shapes specified by sketches while maintaining diversity and realism.
\noindent \textbf{Incremental Learning:} Incremental Learning (IL) \cite{polikar2001learn++, kuzborskij2013n} is a machine learning paradigm where a model adapts itself to learn new tasks sequentially while retaining the previously learnt knowledge. Although deep networks have demonstrated incredible achievements in a variety of tasks \cite{santoro2016meta, snell2017prototypical}, sequentially learning different tasks remains a key challenge. Consequently, IL continues to receive considerable research attention \cite{hsu2018re, icart2017, chaudhry2018efficient, kirkpatrick2017overcoming, aljundi2018memory}. Majority of the present research either use memory-based \cite{hsu2018re, icart2017}, distillation-based \cite{cheraghian2021semantic, dong2021few}, or regularisation-based \cite{kirkpatrick2017overcoming} approaches to tackle the IL task. Based on the task at hand, IL can be categorised into (a) Incremental domain learning \cite{rosenfeld2018incremental}, which aims at performing incremental domain adaptation. (b) Incremental task learning \cite{aljundi2018memory}, where each task consists of separate classification layers, and a task descriptor selects the appropriate layer during the testing phase. (c) Class incremental learning (CIL), the most challenging IL task that operates in a single-head setup with no available task descriptors. In CIL, the model needs to learn a unified classifier to fit all the new unseen classes incrementally. Distillation \cite{li2017learning} and memory-based \cite{hsu2018re} methods are more effective than regularisation-based ones \cite{kirkpatrick2017overcoming} in the CIL setting. This paper is mainly concerned with CIL setup, which is the most challenging task among its variants.

\noindent \textbf{Few-Shot Class-Incremental Learning (FSCIL):} Few shot learning ({{FSL}}) aims at adapting a trained model to learn patterns from novel classes (unseen during training) using only a \emph{few labelled} samples \cite{wang2020generalizing}. Recently, it has experienced rapid proliferation \cite{rezende2016one, snell2017prototypical, vinyals2016matching} in the research community. There are three major swim lanes of the FSL problem: (a) recurrent-based \cite{rezende2016one, santoro2016meta} (b) optimisation-based \cite{rusu2018meta, vuorio2019multimodal}, and (c) metric-based frameworks \cite{gidaris2018dynamic, koch2015siamese}. Our work falls under metric-based methods in which similarity is drawn between the query sample and the novel support classes. Conventional CIL presumes that the incrementally provided novel classes have access to a substantial amount of labelled data. Although in the FSCIL paradigm \cite{tao2020few}, the initial dataset contains sufficient training data (base classes), the subsequently provided novel classes contain only a few labelled samples. Very few methods are present  to tackle the FSCIL problem like, pseudo incremental learning \cite{zhang2021few}, knowledge distillation \cite{cheraghian2021semantic,dong2021few}, neural-gas network \cite{tao2020few}. While existing works intend to build a model to incrementally learn novel classes, we aim at building a model for a much harder and practically applicable sketch-based FSCIL setting that addresses user's privacy concerns.

\keypoint{Minimising Domain Discrepancy:} Minimising sketch-photo domain discrepancy \cite{dey2019doodle} is the key in our problem setup. In this context, the two most relevant branch of literature involves Domain Adaptation (DA) \cite{ganin2015unsupervised} and Domain Generalisation (DG) \cite{li2019episodic, li2018learning}. While DA intends to adapt a model trained on a source domain to perform well on a new target domain using only unlabelled images, the aim of DG is to generalise a model from a set of \textit{seen} domain samples to \textit{unseen} domain samples without accessing the unseen domain instances. Our objective is more aligned with DG as we do not update the model parameters during inference. In this work, we take inspiration from the recent developments \cite{yu2020gradient, mansilla2021domain} in DG to learn a domain-agnostic network, minimising the domain gap between sketch and photo.

% and our objective remains to learn a domain agnostic feature representation minimising the domain gap between sketch and photo input. 
% SUBHO - DONE UNTIL HERE 02-11-2021 0338HRS

\vspace{-0.3cm}
\section{Sketch for Incremental Learning}
\vspace{-0.1cm} 

\subsection{Problem Definition}
\vspace{-0.2cm} 
% \keypoint{Overview}

\keypoint{Dataset:} In few-shot class-incremental learning, we are given with $K_{b}$ \emph{base} classes and $K_{n}$ \emph{novel} classes respectively. From the set of base classes, we have \emph{sufficient} access to labelled samples from \emph{photo} $\mathcal{D}_{base}^P=\{(p_i, {{y_i^p}} )\}_{i=1}^{N_b^s}$   and \emph{sketch}  $\mathcal{D}_{base}^S = \{(s_i, {{y_i^s}} )\}_{i=1}^{N_b^s}$ domains, where $y_i \in \mathcal{C}_{base} = \{C_1^b, C_2^b, \cdots, C_{K_b}^b\}$. On the other side, for novel classes, we have \emph{minimal} access to labelled samples from \emph{only sketch} domain $\mathcal{D}_{novel}^S = \{(s_j, y_j)\}_{j=1}^{N_n^s}$ where number of samples for each novel category is limited, and $y_j \in \mathcal{C}_{novel} = \{C_1^n, C_2^n, \cdots, C_{K_n}^n\}$. Here, base and novel classes are completely disjoint, so that $\mathcal{C}_{base} \cap\mathcal{C}_{novel} = \Phi$.

% {\color{red}We have a classifier $f: x \rightarrow y$ to learn which can be decomposed into two sequential modules. \cut{{ $\mathcal{D}_{base}^S=\{(s_i, y_i)\}_{i=1}^{N_b^s}$}}} 

\keypoint{Model:} We have a neural network classifier, comprising of a  feature extractor $\mathcal{F}_{\theta}$ followed by linear classifier $\mathcal{R}_{w}$, such that $y = \mathcal{R}_{w}(\mathcal{F}_{\theta}(x))$. $\mathcal{F}_{\theta}$ is employed using a convolutional neural network followed by global-average pooling, and given an input image $x \in \mathbb{R}^{h \times w \times 3}$, we get a feature representation as $f_d = \mathcal{F}_{\theta}(x) \in \mathbb{R}^d$. Following \cite{gidaris2018dynamic}, for better generalisation $\mathcal{R}_{w}$ is devised as a \emph{cosine similarity}  function (unlike dot product based typical linear classifier), consisting a learnable $W$ matrix whose size is of $\mathbb{R}^{|C|\times d}$, where $|C|$ is the number of classes. Thus, $\mathcal{R}_{w}: \mathbb{R}^d \rightarrow \mathbb{R}^{|C|}$ outputs a probability distribution over classes as $p(\bar{y}) = {\texttt{softmax}( \hat{W}\cdot \frac{f_d}{{\left \| f_d \right \|_2}})}$. ${\hat{W}}$ is obtained by $l_2$ normalising every $d$ dimensional row-vector ${w}_k \in {W}$ that depicts weight-vector for $k^{th}$ class, \emph{i.e.} $\hat{w}_k =\frac{w_k}{{\left \| w_k \right \|_2}}$.

\begin{figure*}[!hbt]
\includegraphics[width=0.95\linewidth]{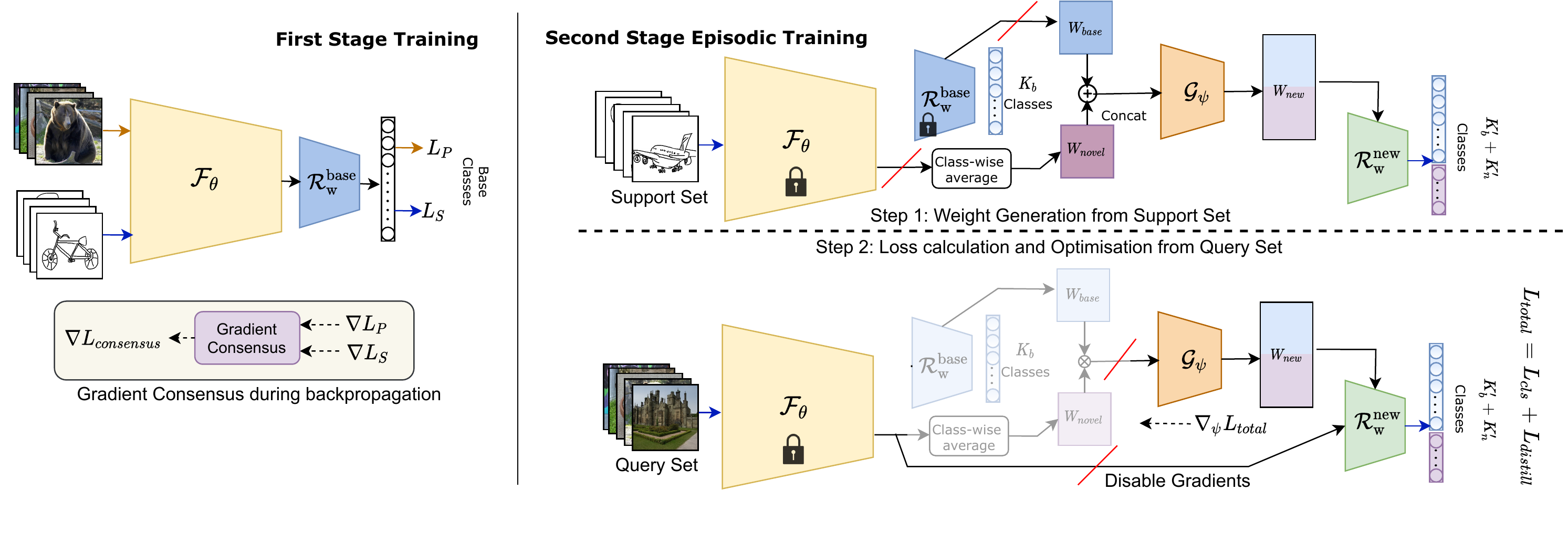}
\vspace{-0.5cm}
\caption{(a) Primarily, we aim to learn a domain-agnostic backbone feature extractor ($\mathcal{F}_{\theta}$) through gradient consensus. (b) In the second stage, we learn a weight generator ($G_{\psi}$) through episodic pseudo incremental learning involving two steps. Firstly, to obtain an updated [base+novel] classifier, a sketch support set is utilised to produce weight vectors for novel classes as well as to refine weight vectors for base classes. Secondly, for loss computation, the resulting weight vectors are evaluated against real photos from both [base+novel] classes.}
\vspace{-0.5cm}
\label{fig_arch}
\end{figure*} 
% During inference, we incrementally learn the new classes with the support of few imaginative sketches doodled by users. 

\keypoint{Learning Objective:} The neural network classifier $\{\mathcal{F}_{\theta}, \mathcal{R}_{w}\}$ is trained from the abundant labelled samples of $K_b$ base classes, and let the initial base classifier be $\mathcal{R}_{w}^{base}: \mathbb{R}^{d} \rightarrow \mathbb{R}^{K_{b}}$ whose weight matrix is $W_{base} \in \mathbb{R}^{K_{b} \times d}$. During inference under FSCIL \cite{tao2020few}, we do \emph{not have} any access to labelled data of base classes, and given only $k$ (small number) \emph{sketch} samples for each of $\mathcal{C}_{n}$ novel categories, we intend to update the classifier  $\mathcal{R}_{w}^{base}$ to $\mathcal{R}_{w}^{new}$ which can recognise \emph{photos} from both $\mathcal{C}_{base} \cup \mathcal{C}_{novel} $ classes. To do so, we need to compute a new weight matrix ${W_{new}} \in \mathbb{R}^{(K_{b}+K_{n}) \times d}$ with respect to $\mathcal{R}_{w}^{new}: \mathbb{R}^{d} \rightarrow \mathbb{R}^{(K_{b}+K_{n})}$ that can perform $(K_{b}+K_{n})${-way} class classification. 

% a few supporting examples from novel classes in terms of \emph{hand-drawn sketches} such that

Therefore, our objective is to figure out a new ${W_{new}}$ matrix for classifier $\mathcal{R}_{w}^{new}$ using the previous base classes' knowledge $W_{base}$  and a few hand-drawn \emph{sketch exemplars} from novel classes such that (i) the knowledge of base classes is not forgotten (preserved), as well as (ii) it quickly adapts to novel classes using few samples, (iii) thus, enabling it to perform well on real photos minimising the domain gap \cite{li2018learning} with sketch samples from novel classes as support. 
\cut{Please note that the ultimate goal of the work is to learn a model, which not only classifies photos from base classes, but also can classify photos from novel classes with support from user-drawn \emph{sketches as exemplar} for unseen classes during few-shot class incremental learning.}
Overall, our framework consists of \emph{three} modules {(i)} a backbone \emph{feature extractor} $\mathcal{F}_\theta$, {(ii)} a \emph{classifier} $\mathcal{R}_w$ {(iii)} a \emph{weight generator} $G_{\psi}$ that will take previous base classifier weights $W_{base}$ and sketch exemplars (support set) from novel classes as input, to generate a new weight matrix $W_{new}$ for updated classifier $\mathcal{R}_{w}^{base} \rightarrow 
\mathcal{R}_{w}^{new}$ in order to classify real photos from \emph{both} base and novel classes. 

\vspace{-0.1cm} 
\subsection{Cross Modal Pre-Training for Base Classes}
\vspace{-0.1cm} 
Our framework follows a \emph{two}-stage training. In the first stage, we train the model for base classes using standard cross-entropy loss, while in the second stage, we learn the \emph{weight generator} via few-shot pseudo-incremental learning. \cut{We assume that we have sufficient access to labelled data from base classes.} Once trained, we freeze the weights of $\mathcal{F}_{\theta}$ in the next stage to (i) avoid over-fitting during the few-shot update and (ii) to alleviate catastrophic forgetting \cite{kirkpatrick2017overcoming} of the base classes. 

% \textcolor{red}{
% In comparison to existing few-shot incremental learning, we need to address the domain gap between photos and sketches, so that we can categorise new class images in the real photo domain by using a common feature extractor and the incremental class knowledge obtained through sketch exemplars (support set). \textcolor{blue}{(or)}  We need to overcome the domain gap between photos and sketches in comparison to existing few-shot incremental learning, so that we can categorise new class images in the real photo domain with ease by utilising a common feature extractor and the incremental class knowledge acquired through sketch exemplars (support set). 
% }

% Unlike existing few-shot incremental learning, we need to handle the domain gap between photos and sketches, so that we can use a common feature extractor and knowledge of incremental classes acquired through sketch exemplars to classify novel class images in real photo domain.

Unlike existing few-shot incremental learning, we need to handle the domain gap \cite{li2019episodic} between photos and sketches, so that the knowledge of incremental classes acquired through sketch exemplars can classify novel class images in real photo domain.
As we have sufficient access to labelled training data from both photo and sketch domains for base classes, a very straightforward way to handle the domain gap is to train by combining labelled photos and sketches (spatially extended) with equal probability in every mini-batch -- so that the model generalises equally well on both photos and sketches. Given an input $x$, let the model's output be $\bar{y} = \mathcal{R}_{w}^{base}(\mathcal{F}_{\theta}(x))$ where labelled data $(x, y)$ come from either photo $(p,{{y^p}}) \sim \mathcal{D}_{base}^P$ or sketch $(s,{{y^s}}) \sim \mathcal{D}_{base}^S$ domain with $x \in \{p, s\}$ and $y$ ({{$y^s$ or $y^p$}}) being corresponding one-hot encoded class label. The cross-entropy loss  $\mathcal{H}(\cdot,\cdot)$ can be calculated as $\mathcal{L} = \mathcal{H}(\bar{y},y) = \sum_{i=1}^{K_{b}} y_i \log p(\bar{y}_i)$. Against a batch having $b$ photos and $b$ sketches, we can calculate the individual loss across photos and sketches as $\mathcal{L}_{P}$ and $\mathcal{L}_{S}$, respectively. Thereafter, we update the model by taking gradient $\nabla  \mathcal{L}_{total}$ over total loss which is given as follows:  
\vspace{-0.2cm}
\begin{equation}
    \mathcal{L}_{total} = \frac{1}{b} \hspace{-0.5cm} \sum_{(p,y)  \sim \mathcal{D}_{base}^P} \hspace{-0.5cm} \mathcal{L}_{P}(p,y) +  \frac{1}{b} \hspace{-0.5cm} \sum_{(s,y)  \sim \mathcal{D}_{base}^S} \hspace{-0.5cm} \mathcal{L}_{S}(s,y)
    \vspace{-0.25cm}
\end{equation}
% \vspace{-0.25cm}

% \begin{multline}
%     \mathcal{L}_{total} = \mathcal{L}_{s} + \mathcal{L}_{p} \\
%                       = \hspace{-0.5cm} \sum_{ (p,y)  \sim \mathcal{D}_{base}^P} \hspace{-0.5cm} \mathcal{H}\Big(\mathcal{R}_{w}^{base}(\mathcal{F}_{\theta}(p)), y\Big) + \hspace{-0.5cm} \sum_{(s,y)  \sim \mathcal{D}_{base}^S} \hspace{-0.5cm} \mathcal{H}\Big(\mathcal{R}_{w}^{base}(\mathcal{F}_{\theta}(s)),y \Big)
% \end{multline}

However, naively training with two significantly different domains (photo vs sketch) gives rise to \emph{conflicting gradients} within each batch, as information specific to the one domain might be irrelevant to the other, thereby suppressing the generalisation capability of the model. In other words, the information carried by $\nabla  \mathcal{L}_{P}$ and $\nabla  \mathcal{L}_{S}$ might not mutually agree, and adding them naively would lead to inhibiting \cite{yu2020gradient} the training signal overall.

\keypoint{Gradient Consensus:} Inspired from multi-task learning \cite{yu2020gradient} and domain generalisation \cite{mansilla2021domain} literature, we aim to update the model in the direction where there is an agreement in the gradient space between two domains in order to learn a domain invariant representation. In particular,  gradient vectors having the same sign will be retained, while those having conflicting signs will be set to \emph{zero}, as shown in Eq. \ref{gradsurgery}. Here, the $\mathrm{sig}(\cdot)$ is a sign operator, and $\nabla L_{P}^n$ and $\nabla L_{S}^n$ denote the $n$-th component of the gradient associated to photo and sketch domain respectively. The gradient consensus function $\delta(\cdot, \cdot)$ checks element-wise if the signs of the gradient components match, and it returns 1 if all components have the same sign for a given $n$; otherwise 0.

\vspace{-0.6cm}
\begin{equation}\label{gradsurgery}
 \delta (\nabla L_{P}^n, \nabla L_{S}^n) =\left\{\begin{array}{lll}
                1, &  \mathrm{sig}(\nabla L_{P}^n)= \mathrm{sig}(\nabla L_{S}^n) \\
                0, &  \textnormal{otherwise}
            \end{array}\right.
\end{equation} 
\vspace{-0.1cm}
\vspace{-0.1cm}
\begin{equation}\label{consensus_grad}
\nabla L_{consensus}^{n} =\left\{\begin{array}{lll}
                \nabla L_{P}^{n} + \nabla L_{S}^{n}, & \textnormal{if} &  \delta^n =1 \\
                0, &  \textnormal{if} &  \delta^n=0
            \end{array}\right.
\vspace{-0.1cm}
\end{equation}

\cut{Here, we set the sign-conflicting gradient vectors to zero and sum the gradient vectors when there is a sign-agreement.} This gradient agreement strategy helps to reduce the harmful cross-domain gradient interference while updating the model parameters using $\nabla L_{consensus}^{n}$. Thus, enabling us to adjust the model parameters in a direction that helps to improve generalisation across both sketch and photo.

 % we can normalise the gradients to make it different from the gradient surgery paper. 

% Exemplars - support set belonging to its class
\vspace{-0.1cm}
\subsection{Few-Shot Classifier Weight Generation}
\vspace{-0.2cm}
% Give brief intro: 

% WGS
% WRS
% Weight Generation strategy
% Weight Refinement mechanism/strategy 

%SUBHO
\keypoint{Overview:} In order to classify photo from novel classes, we need to design a mechanism that can generate \emph{additional} weight vectors for the novel classes. 
% Weight vectors  of the novel classes are essential to classify those classes. Consequently, it is instrumental to design a mechanism, which we can generate \emph{additional} weight vectors for the novel classes. 
As we assume that only a few supporting hand-drawn sketch exemplars will be provided corresponding to every novel class, we design the weight generator $G_{\psi}$ under a few-shot paradigm \cite{wang2020generalizing}. $G_{\psi}$ produces weight vectors for novel classes and also re-generates (refines) weight vectors for base classes in order to get a better overall decision boundary in the presence of novel classes. Here, the two major objectives are (i) learn the knowledge of novel classes from \emph{fewer sketch exemplars}, while \emph{classifying photos} of novel classes through cross-modal generalisation (ii) \emph{not to degrade} the performance of base classes while learning the novel ones.

% To accomplish so, we create a weight generation module $G_{\psi}$ that produces weight vectors for novel classes and also re-generates (refines) weight vectors for base classes in order to get a better overall decision boundary in the presence of novel classes.  As we assume that only a few supporting hand-drawn sketch exemplars will be provided corresponding to every novel class, we design the weight generator under a few-shot paradigm \cite{wang2020generalizing} with two major objectives (i) learn the knowledge of novel classes from \emph{fewer sketch exemplars}, while \emph{classifying photos} of novel classes through cross-modal generalisation (ii) \emph{not to degrade} the performance of base classes while learning the novel ones.

We employ sketch exemplars as a \emph{support set} to generate the \emph{new} weight matrix following the episodic training \cite{li2019episodic} of few-shot learning. To determine the loss for updating the weight generating module, the quality of the generated weight matrix is assessed against a \emph{query set} of photo samples. In particular, there are \emph{two} steps \cite{snell2017prototypical} while training the weight generation module. (i) {\emph{Weight generation} using support set:} \cut{(\textcolor{red}{weight generation scheme/mechanism/strategy WGS}) (\textcolor{red}{weight refinement scheme/mechanism/strategy WRS})} sketch exemplars as support set are used together with $W_{base}$ to generate the new weight matrix $W_{new}$ (comprising both base and novel classes) (ii) {\emph{Loss calculation} on query set:}  $W_{new}$ is used to classify query set \emph{photos} in order to calculate loss, which is then utilised to optimise the weight generation module using gradient descent.

\keypoint{Weight Generation:} $G_{\psi}$ takes two things as input (i) $W_{base}$ from $\mathcal{R}_{w}^{base}$ representing the knowledge of previous base classes (ii) class-wise representative features of novel classes from sketch exemplars. We assume to have access to $k$ sketch samples for each of the $K_{n}$ novel classes -- the support set. A straightforward way to get class-wise representative vectors is to average feature representations of sketches for each individual classes. In particular, for $j^{th}$ novel class, the representative vector can be calculated as:
\vspace{-0.4cm}
\begin{equation}
    w_{j}^{novel} = \frac{1}{k}\sum_{i=1}^{k} \mathcal{F}_{\theta}(s_i)
\vspace{-0.3cm}
\end{equation}

Thereafter, by applying $l_2$ norm on each $ w_{j}^{novel}$, we can naively form the weight vectors of novel classes as $W_{novel} = \{w_{1}^{novel}, w_{2}^{novel}, \cdots, w_{K_{n}}^{novel}\} \in \mathbb{R}^{K_n \times d}$. The easiest way for incremental learning would be to use naive concatenation to get new weight matrix as $[W_{base}; W_{novel}] \in \mathbb{R}^{(K_{b} + K_{n}) \times d}$. However, it has \emph{two} major limitations (i) $W_{novel}$ remains unaware about the knowledge of bases classes (ii) $W_{base}$ which was discriminative across the base classes might lose its representation-potential when we add additional weight vectors of novel classes without modelling a mutual agreement strategy for learning discriminative decision boundaries across all $K_{b} + K_{n}$ classes. Thus, to attain an optimal decision boundary for all classes under incremental setup, an \emph{information passing} mechanism is critical for $W_{new}$ generation. 

% the global context information from other classes for $W_{new}$ generation.

% Thus, to attain an optimal decision boundary for all classes under incremental setup, it is important to consider the global context information from other classes for $W_{new}$ generation.

\keypoint{Message Passing:} For \emph{information-propagation} among weight vectors of $K_{b}+K_{n}$ classes, we use Graph Attention Network (GAT) \cite{velivckovic2017graph}. GAT is a good choice for information-propagation owing to its permutation-invariance to sequence of weight vectors \doublecheck{as the novel classes may appear in any order}. As the weights are shared across different nodes, it can also handle incoming variable number of novel classes effortlessly. The input to GAT is given as $W_I = \{w_{1}^{base}, \cdots, w_{K_{b}}^{base}, w_{1}^{novel}, \cdots, w_{K_{n}}^{novel}\}$ having $K_{total} = K_{b} + K_{n}$ weight vectors, where each $w_i \in \mathbf{R}^d$ denotes an input to a specific node to GAT. First it computes relation co-efficient between every pair of node by inner product operation as $e_{i,j} =  \left \langle V_aw_i, V_bw_j \right \rangle$, with two learnable linear embedding weights $V_a$ and $V_b$. $e_{i,j}$ is\cut{ again} normalised by softmax function to get the attention weights with respect to node $i$ as: $a_{ij} = \frac{\exp(e_{ij})}{\sum_{k=1}^{K_{total}} \exp(e_{ik})}$. The update rule for $i^{th}$ node gathering information from all other nodes becomes  
\vspace{-0.4cm}
\begin{equation}
    w_{i}^{update} = w_{i} + \big(\sum_{j=1}^{K_{total}}a_{i,j} V_c w_{i}\big)
    \vspace{-0.3cm}
\end{equation}
\noindent where, $V_c$ is a learnable linear transformation. We repeatedly update the weight vectors at every node in the graph, and finally we obtain the generated weight vectors for both base and novel classes as $W_{new}$. \cut{ \in \mathbb{R}^{(K_{b} + K_{n}) \times d}.}
 In brief, $W_{new} = G_\psi(W_I): \mathbb{R}^{(K_b+K_n) \times d} \rightarrow \mathbb{R}^{(K_b+K_n) \times d}$, where $W_I = [W_{base}; W_{novel}] \in \mathbb{R}^{(K_b+K_n) \times d}$, thus we generate the weight vectors for both base and novel classes during incremental learning.
% global contex information of individual classes

% $w_{j}^{novel} = \frac{1}{k}\sum_{i=1}^{k} \mathcal{F}_{\theta}(s_i)$ . 

% by episodically
% constructing pseudo incremental tasks based on the
% base dataset D0
% train to mimic the test scenario.

% % takes inspirations from meta-learning [42], where
% a small classification task is constructed to enable learning
% on the meta-level beyond a specific task.

% Multi-Modal Episodic based Incremental training to minimise domain discrepancy(for domain generalisation) 
\keypoint{Episodic Pseudo Incremental Training:}  
Keeping the feature extractor $\mathcal{F}_\theta$ fixed, we train the few-shot weight generator $G_\psi$ taking inspiration from \cut{the recent developments of} few-shot learning literature \cite{snell2017prototypical, rezende2016one, santoro2016meta}. 
% Once the first stage model is trained from base classes, we fix the backbone feature extractor $\mathcal{F}_\theta$, and only train  few-shot weight generator $G_\psi$ in the second stage. To train the few-shot weight generator, we take inspiration from the recent developments of few-shot learning \cite{snell2017prototypical} and meta-learning \cite{li2019episodic}. 
As the training dataset is limited, we episodically construct pseudo incremental task based \emph{only} on the base classes to mimic the real testing scenario. 
 
In particular, following the first stage of training, we get classifier weight matrix of base classes as $W_{base} \in \mathbb{R}^{K_{b} \times d}$.
In order to create each episode, we \emph{synthetically drop} $K_{n}'$ weight vectors from $W_{base}$, and we treat those corresponding classes as pseudo novel classes whose weights now need to be generated. That means, at a particular episode, the pseudo base class matrix becomes $W_{base}' \in \mathbb{R}^{K_{b}' \times d}$ where $K_{b}' = K_{b} - K_{n}'$. Thereafter, corresponding to those \emph{dropped} base classes which now become pseudo novel classes, we use $k$ sketch samples for each of the pseudo novel classes as the \emph{support set} to first generate representative class-wise weight vectors $W_{novel}' \in \mathbb{R}^{K_{n}' \times d}$, which is again fed to GAT together with $W_{base}'$ for relationship modelling to generate pseudo $W_{new}'$. In every episode, while support set ($\mathcal{S}$) is used to generate the classifier weights, another query set ($\mathcal{Q}$) involving real photos from both pseudo base and novel classes are fed through pre-trained backbone followed by classifier with newly generated weight matrix $W_{new}'$ to compute loss for optimisation. Please refer to Fig.~\ref{fig_arch}.

In contrast to earlier FSCIL works \cite{tao2020few, dong2021few}, our episodic training is cross-modal in nature, where the support and query set consist of sketch and photo respectively. As training is done over base classes with pseudo-novel classes, we found mixing both sketch and photo in the support set with \emph{gradient consensus} generalises better on real photos. However, sketch acts as the only exemplars during real inference. 

% Empirically, we found  mixing both sketch and photo in the support set with \emph{gradient consensus} helps to generalise better on real photos during final classification under real testing scenario.
% \textcolor{red}{here we also use gradient surgery to minimise the domain gap. This part is missing. Here we adapt to sketches. }
% there are two steps during training the weight generation module (i) \emph{weight generation:} use sketch exemplars as support set and $W_{base}$ to generate the new weight matrix $W_{new}$ (comprising both base and novel classes) (ii) \emph{loss calculation on query set:} use $W_{new}$ to classify \emph{photos} from query set to calculate loss through which weight generation module is finally updated by gradient descent. 

\keypoint{Loss Functions:} Contrary to fully supervised classification from abundant training data, few-shot learning \cite{snell2017prototypical} is more challenging as only a few samples are available for the new weight matrix generation. Given this rationale, we aim to design the pseudo incremental learning by dropping weights vectors from $W_{base}$, which is learned from base classes through standard supervised classification. We aim to see if the fully supervised knowledge learned in $W_{base}$ could provide training signal \cite{hinton2015distilling} to learn the $G_{\psi}$.\cut{Furthermore, $W_{base}$ learned through gradient agreement strategy could provide the domain invariant learning signal so that the domain discrepancy between support sketch and query photos could be minimised.}To \cut{accomplish this}do so, we additionally define a \emph{distillation loss}  along with standard \emph{classification loss} calculated over the query set, which acts a consistency regularisation. \doublecheck{\cut{Consistency regularisation}This ensures that weight vectors predicted by the weight generator remain close to what has been learned through supervised classification from first stage.} In particular, following few-shot weight generation we get an incrementally learned classifier $\mathcal{R}_{w}^{new}$ with generated weight matrix $W_{new}$. On the other side we already have $\mathcal{R}_{w}^{base}$ learned from first stage pre-training. Given a photo $p$ from query set ($\mathcal{Q}$), for distillation loss we treat the \emph{soft} prediction using $\mathcal{R}_{w}^{base}$ as a ground-truth to calculate the distillation loss. Thus, the total loss becomes $\mathcal{L}_{total} = \mathcal{L}_{cls} + \mathcal{L}_{distil}$ which is used to train $G_\psi$. \cut{Given}If $\mathcal{H}(\cdot, \cdot)$ be cross-entropy loss, $\mathcal{L}_{cls}$ and $\mathcal{L}_{distil}$ are defined as:

% \textcolor{red}{ Learning from few samples is a challenging task (that too from cross domain). So additional loss is required to generate weight vectors with consistence(without nuance). We can call it as Consistency training instead of knowledge distillation. Consistency regularisation. Enforces to learn without much deviation. ) } 

\vspace{-0.3cm}
\begin{equation}
 \mathcal{L}_{cls}  =  \frac{1}{ \left | \mathcal{Q} \right |}  \hspace{-0.1cm}   \sum_{(p,y)  \sim \mathcal{Q}} \hspace{-0.25cm} \mathcal{H}(\mathcal{R}_w^{new}({\mathcal{F}_{\theta}(p)}), y)
\end{equation}
\vspace{-0.25cm}
\begin{equation}
\mathcal{L}_{distil} = \frac{1}{ \left | \mathcal{Q} \right |} \hspace{-0.2cm}  \sum_{(p,y)  \sim \mathcal{Q}} \hspace{-0.2cm} \mathcal{H}(\mathcal{R}_w^{new}({\mathcal{F}_{\theta}(p)}), \mathcal{R}_w^{base}({\mathcal{F}_{\theta}(p)}))
\vspace{-0.2cm}
\end{equation}

\cut{
\setlength{\tabcolsep}{6pt}
\begin{table*}[h]
    \centering
    \footnotesize
    \caption{Accuracy}
    \begin{tabular}{cccc}
        \hline \hline
        Method & Acc Both & Acc Base & Acc Novel \\
        \hline
         Photos & 60.48\% & 64.75\% & 78.48\% \\
         \hdashline
         GNN + Exemplar aug & 49.06\% & 61.88 & 63.8 \\
         GNN + no aug & 48.38 & 63.47 & 65.152 \\
         GNN + Exemplar weak aug + grad sur & 49.60 & 62.6107 & 64.5156 \\
         GNN + attention & 50.16 & 61.02 &  64.63 \\
         feat align & 45.71 & 65.57 & 62.94 \\
         feat align + wass & 46.4276 & 64.7409 & 61.5769 \\
         rkd & 47.4162 & 62.9231 & 60.0956 \\
         GNN + rkd + clip5 & 46.8009 & 63.970 & 65.0147 \\
         feat_avg + wass & 45.2436 &  63.736 & 62.6369 \\
         KD & 49.09 & 63.82 & 63.21 \\
         GNN + photo-sketch Exemplar(during train) + grad surgery &  51.1529 & 64.9902 & 64.1662 \\
         GNN + photo-sketch Exemplar(during train) + no grad surgery &  50.15 & 65.68 & 63.51\\
         GNN + photo-sketch Exemplar(during train) + grad surgery(best novel) &  49.72 & 65.22 & 64.91 \\
         GNN + photo-sketch Exemplar(during train) + no grad surgery(best novel) & 49.27 & 65.02 & 64.93 \\
         Baseline SB-FSCIL \\
         Prototypical SB-FSCIL \\
         PB-FSCIL \\
         % Active learning GCNN +  photo-sketch Exemplar(during train)  + KD & 49.49 & 64.03 & 61.14 \\
         
    \end{tabular}
    \label{tab:main_results}
\vspace{-0.3cm}
\end{table*}
}

\vspace{-0.2cm}
\section{Experiments}
\vspace{-0.2cm}
\keypoint{Datasets:} We evaluate our DIY-FSCIL framework on the popular Sketchy dataset \cite{sangkloy2016sketchy} which is a large collection of photo-sketch pairs. As paired photo-sketch is \emph{not essential} for our framework, we use the extended version of Sketchy with $60,502$ additional photos that Liu \etal \cite{liu2017deep} later introduced for category-level SBIR. In particular, Sketchy-extended comprises 125 categories with $75,471$ sketches and $73,002$ images in total. Existing zero-shot SBIR \cite{dey2019doodle, dutta2019semantically} works split the dataset into $104/21$ disjoint classes for training/testing(unseen). We keep the same $21$ classes for testing (novel classes), while for hyperparameter tuning, out of 104 classes, we consider $64$ for training and the rest $40$ classes for validation. In summary, we call them $T_{train}$ (64 classes), $T_{val}$ (40 classes), and $T_{test}$ (21 classes) respectively. The train set ($T_{train}$) is often referred to as \emph{base dataset} and is further split into three subsets $(T_{train}^{train}:T_{train}^{val}:T_{train}^{test})=(60\%:20\%:20\%)$. The subset $T_{train}^{test}$ is used to evaluate the overall performance on the base classes during incremental setup. The steps outlined above are followed for both sketches and photos. For every model evaluations, we follow the same settings, including the categories' division and incremental training samples. 
% For hyperparameter tuning, out of 104 classes, we consider    

\setlength{\tabcolsep}{6pt}
\begin{table*}[t]
\centering
\caption{\doublecheck{Average classification accuracy of DIY-FSCIL framework using our self-designed baselines and adopted SOTA FSCIL \cite{gidaris2018dynamic} (not specifically designed for cross-modal). For every experiment, we create $600$ episodes each with $5$ random classes from both novel and base categories separately.  Each episode contains a total $15\times5$ (15 samples from each of the 5 classes) and $15\times5$ query photos from from both novel and base categories respectively. $B5^*$ is an upper bound.}}    
    
% \caption{Classification accuracy under DIY-FSCIL setup using self-designed baselines and SOTA FSCIL (not designed for cross-modal). For every experiment, we create $600$ episodes each with $5$ random classes from both novel and base categories separately. 
% Each episode contains a total $15\times5$ (15 samples from each of the 5 classes) and $15\times5$ query photos from both novel and base categories respectively.}
% and different ablated versions of our framework
    \vspace{0.05cm}
    \footnotesize
    \begin{tabular}{ccccccccccc}
        \hline
        \multicolumn{3}{c}{\multirow{2}{*}{Methods}}
        &   \multicolumn{3}{c}{5-Shot Learning} & \multicolumn{3}{c}{1-Shot Learning} \\
        \cline{4-6}  \cline{7-11}
         & & & Acc@both & Acc@base &Acc@novel & Acc@both & Acc@base &  Acc@novel\\
        \hline
        \multirow{5}{*}{Baselines} 
         & \multicolumn{2}{c}{$B1$} & 36.29 \% & 73.94\% & 38.92\% & 31.52\% & 73.98\% & 34.68\%\\
         & \multicolumn{2}{c}{$B2$} & 25.86\% & 32.85\% & 70.58\% &  28.81\% & 40.91\% & 50.24\%\\
        
        & \multicolumn{2}{c}{$B3$} & 58.92\% & 73.81\%  & 72.34\% & 53.35\% & 73.75\%  & 59.93\%\\
        
        & \multicolumn{2}{c}{$B4$} &  54.5\% & 71.68\% & 71.81\% & 51.41\%  & 71.68\% & 51.44\%\\
        
        & \multicolumn{2}{c}{$B5^*$} & 71.52\% & 75.72\% & 85.46\% &  63.47\% & 75.83\% & 73.90\%\\
        \hdashline
         \multirow{3}{*}{SOTA FSCIL} 
         & \multicolumn{2}{c}{\cite{gidaris2018dynamic}} &  50.45\% & 74.35\% & 65.81\% &  44.71\% & 73.98\% & 64.21\% \\
        & \multicolumn{2}{c}{\cite{snell2017prototypical}} & 45.25\% & 74.10\% & 63.46\% & 41.97\% & 74.60\% & 61.85\%\\
        & \multicolumn{2}{c}{\cite{tao2020few}} & 51.54\% & 73.21\% & 66.82\% & 45.81\% & 73.58\% & 63.95\%\\
  \hdashline
         Ours 
         
        & \multicolumn{2}{c}{DIY-FSCIL} & 60.54\% & 74.38\% & 75.84\% & 54.97 \% & 74.06\% & 64.10\%\\
       \hline
    \end{tabular}
    \label{tab:maintable2}
        \vspace{-0.3cm}
\end{table*}

\keypoint{Implementation Details:} We have implemented the DIY-FSCIL framework using PyTorch \cite{paszke2017automatic} and conducted the experiments using one \doublecheck{11-GB NVIDIA RTX 2080-Ti GPU}. We employ the standard ResNet18 model  as the backbone feature extractor ($\mathcal{F}_{\theta}$).  The features of the input image are derived from the final pooling layer of the $\mathcal{F}_{\theta}$ with a dimension of $d=512$. We use a one-layer GAT to design our weight generator $G_{\psi}$. In the initial stage, the feature extractor ($\mathcal{F}_{\theta}$) is trained on the training set $T_{train}^{train}$. We train the $\mathcal{F}_{\theta}$ for 100 epochs, and during the second stage $\mathcal{F}_{\theta}$ is freezed and the weight generation module involving GAT is trained for 60 epochs. We use SGD optimiser with learning rate 0.01 and batch size of 8 for all experiments. In order to reduce the error caused by the random sampling of the incremental classes and its samples, we report the average results obtained by \emph{five} different seeds. 

% The exemplars embeddings are averaged in the  $G_{\psi}$ module to obtain $ w_{j}^{novel}$ ($\mathcal{R}_{w_{j}}^{novel}: \mathbb{R}^{n \times d} \rightarrow \mathbb{R}^{1 \times d}$). All these classification weight vectors are concatenated and passed to a GAT network which accepts  $\mathbb{R}^{n \times d}$ as input and returns a  refined weight vectors of dimension $\mathbb{R}^{n \times d}$. 

%  recap the evaluation process -- as pseduo training and real evaluations are different
\vspace{-0.1cm}
\subsection{Evaluation Protocol}
% \vspace{-0.1cm}
Following the incremental step $\mathcal{R}_{w}^{base} \rightarrow \mathcal{R}_{w}^{new}$, we evaluate the performance of staged operations $\mathcal{F}_{\theta} \circ \mathcal{R}_{w}^{new}$ under three circumstances -- (a) upon only novel classes, (b) upon only base classes, and (c) upon both base plus novel classes. While for only novel classes the class label space consists of $y_j \in \mathcal{C}_{novel} = \{C_1^n \cup  C_2^n  \cdots  C_{K_n}^n\}$, the same for only base classes becomes $y_j \in \mathcal{C}_{base} = \{C_1^b \cup  C_2^b \cdots C_{K_b}^b\}$. Furthermore, for evaluation under base plus novel classes, the label space spans across $y_j \in \mathcal{C}_{both} = \{\mathcal{C}_{base} \cup \mathcal{C}_{novel}\}$. These three evaluating situations answer -- (a) how well the model adapts to novel classes from few (1 or 5) sketch examples, (b) how well the model is able to preserve the accuracy (mitigating catastrophic forgetting) of the base classes for which the training data is inaccessible during the incremental step, (c) how well the model performs overall for both base and novel classes. Following the two-stage training using $T_{train}^{train}$, i.e., pre-training on base-classes followed by learning few-shot weight generator, we obtain $\mathcal{F}_{\theta}$ and $G_{\psi}$, which are used for inference under incremental setup.

\keypoint{Evaluation of novel classes (Acc@{novel}):} \cut{For this setting without base classes, we use $\mathcal{C}_{novel}$.} Test set ($T_{test}$) is used to create few shot tasks similar to episodic training.
% During testing, one would make use of test set ($T_{test}$) to create few shot tasks similar to episodic training. 
These few shot tasks are formed by sampling $K_{novel}=5$ categories. Then, we sample one (1-shot) or five (5-shot) exemplars per category (sketches) and $15$ query samples per category (photos). Here, the query samples will be from the same novel categories but we make sure that they do not overlap with the exemplars under a particular episode. \doublecheck{$G_{\psi}$ uses exemplar embeddings obtained via $\mathcal{F}_{\theta}$, along with base weights, to generate incremented classifier's weight $W_{new}$, which is then evaluated on the query set.} Apart from helping to understand the model's capability to learn novel classes in a few-shot setting excluding the base classes, this metric also helps assessing the \emph{model's generalisation capability} on \emph{cross-domain} data. 
% Average results of $600$ few-shot tasks \cite{gidaris2018dynamic} are reported.
Following the existing FSCIL literature \cite{gidaris2018dynamic}, we create $600$ few-shot tasks  and report the average results from them.

% We cite  the 
% $$ {s_{support}}_{n}^{s} \cap {s_{query}}_{n}^{p} = \Phi$$ (${s_{query}}_{n}^{p}$)
\keypoint{Evaluation of base classes (Acc@{base}):} To verify the potential of mitigating the catastrophic forgetting issue, we evaluate the recognition performance on base categories using the $T_{train}^{test}$ subset on the incremented classifier $\mathcal{R}^{new}_w$. Here, we create few shot tasks by randomly sampling $K_{b}=5$ categories from the base classes without replacement, followed by evaluating with $15$ query photos for each category.

\keypoint{Evaluation of all the classes (Acc@{both}):} Here the label space spans across all the classes ($y_j \in \mathcal{C}_{both}$). In each episode, we sample from all $K_{b}$ base and $K_{n}$ novel classes. Then, we sample one (1-shot) or five (5-shot) exemplars per novel category (sketches), 15 query samples for each of the base and novel categories (photos) to evaluate the performance. This metric helps to determine how the base classes' knowledge affects novel classes and vice-versa. 
% Here the label space becomes $y_j \in \mathcal{C}_{both}$. 

\vspace{-0.1cm}
\subsection{Competitors}
 \vspace{-0.1cm}
% To the best of our knowledge, there exists no prior work dealing with sketch-based FSCIL. Thus, we implement the following set of baselines and their adaptions in order to assess the contribution of our proposed framework.

As there exists no prior work dealing with sketch-based FSCIL, we implement the following set of baselines and their adaptions in order to assess the contribution of our proposed framework. $\bullet$ \textbf{B1:} We use a combination of old-base and new-novel classes to \emph{retrain} the complete model. Besides requiring a lot of computational power, this suffers from a severe class imbalance problem between sufficiently available base classes and few exemplars from novel classes.  Nevertheless, this can not be realised in a real scenario. $\bullet$ \textbf{B2:}  \cut{\textcolor{red}{We keep the backbone feature extractor fixed, and only fine-tune the final classification layer on the novel classes. It acts as a naive baseline, and is limited due to the issue of catastrophic forgetting. }} We only fine-tune the model using the novel classes. It acts as a naive baseline, and is limited due to the issue of catastrophic forgetting.  $\bullet$ \textbf{B3:}  We freeze backbone feature extractor $F_\theta$, and use the class-wise average feature of sketch exemplars as the representative weight-vectors of novel classes along with the pre-trained base-classifier. In other words, we remove the GAT module from our proposed framework. $\bullet$ \textbf{B4:} We further examine the performance of our framework by training both the $\mathcal{F}_{\theta}$ along with the $\mathcal{G}_\psi$. This is used to analyse the importance of freezing the feature extractor $\mathcal{F}_{\theta}$. $\bullet$ \textbf{B5:} During testing, we utilise real images as the support set. As images are more detailed than sketches, this model serves as our upper boundary. However, it fails to address our main concern of violating the data privacy norm. For a fair comparison, we utilise the same settings for all the models as our framework.
\doublecheck{Though existing FSCIL methods \cite{gidaris2018dynamic, tao2020few, snell2017prototypical} are not specifically designed to deal with cross-modal sketch exemplars, we naively adopt those under our sketch-based FSCIL setup.}

\vspace{-0.1cm}
\subsection{Performance Analysis}
\vspace{-0.1cm}
In Table \ref{tab:maintable2}, we report the comparative results using the standard \emph{one}-shot and \emph{five}-shot sketch-based FSCIL setting on Sketchy dataset. We make the following observations: (i) Despite using abundant memory and computational resources \textbf{B1} performs poorly on the novel classes, due to the absence of any mechanism to handle few shot classes (i.e., severe class imbalance). This suggests that few shot paradigm is essential to perform reasonably well on novel classes. (ii)  \textbf{B2} adapts fine-tuning on the novel classes without heavy computational overhead. However, doing so declines the model's performance on the base classes due to catastrophic interference. (iii) While \textbf{B3} outperforms baselines \textbf{B1} and \textbf{B2}, it fails to model mutual agreement between base and novel classes for learning discriminative decision boundaries under incremental setup, revealing the importance of our weight refining strategy through GAT module. (iv) Low performance of \textbf{B4} signifies the necessity of freezing the weights of $\mathcal{F}_{\theta}$ during the second stage of training in order to reduce the catastrophic forgetting problem, and also to generalise notably better on \emph{unseen} categories. (v) \textbf{B5} (upper bound) achieves the best numbers, as the support set comes directly from photos, and this is unlike ours where we have a critical challenge due to the domain gap between sketch exemplars and query photos. (vi) Moreover, the performance of SOTA FSCIL methods is limited by a margin of $9.09\%$ under DIY-FSCIL setup. 

To summarise, our framework helps in solving the challenging DIY-FSCIL problem by both alleviating the catastrophic forgetting of the old classes and enhancing the learning of the new classes under a cross-modal sketch-based few shot setting. Moreover, the proposed framework effectively enables the users to build their own novel classes with the support of their imaginative drawings.

% \vspace{-0.1cm} 
\subsection{Further Analysis and Insights}
\vspace{-0.1cm}
\noindent \textbf{Ablation Study:} We further dive deeper to figure out the contribution of individual design components in Table \ref{tab:ablation}. (i) \emph{GAT:} To access the importance of weight refinement, we remove the GAT module and adapt the framework accordingly. Consequently Acc@novel significantly drops to $62.34\%$ with a decrease of $3.5\%$ for $5$-shot case, and is more pronounced for $1$-shot context, where we perceive larger a drop of $4.17\%$. This observation further strengthens our initial assumption that GAT models an effective mutual agreement strategy for learning discriminative decision boundaries across all the $K_{b} + K_{n}$ classes. (ii) \emph{Gradient Consensus (GC):} The use of GC improves the model's performance substantially, and this is particularly apparent in the initial stages. During the initial stage training, GC improves the model accuracy by $2.72\%$ via effective handling of the harmful cross-domain gradient interference while updating the model parameters.  (iii) \emph{Knowledge Distillation (KD):} Knowledge distillation-based regularisation seeks to provide stability and enforces weight generation module learning. Getting rid of it reduces the Acc@both by a significant $3.75\%$($3.07\%)$ for $1$($5$)-shot setting, thus illustrating its need. (iv) \emph{Cross Modal Training (CMT):} While we use only sketch exemplars as the support set during real inference, during episodic training we mix both sketch and photos along with gradient consensus strategy to bridge the domain gap in weight generation process. Removing this cross-modal training drops the Acc@both by $27.3\%$($25.35\%$) for $1$($5$)-shot setting. In summary, all of the components work in unison to produce the best overall performance.

% \vspace{-0.2cm} 
\setlength{\tabcolsep}{4.5pt}
\begin{table}[t]
    \centering
    \caption{Ablative study: GAT (Graph Attention Network), GC (Gradient Consensus), KD (Knowledge Distillation Loss), CMT (Cross-Modal Training)}
    \scriptsize
    \begin{tabular}{cccccccc}
        \hline
        \multirow{2}{*}{GAT} & \multirow{2}{*}{GC} & \multirow{2}{*}{KD} & \multirow{2}{*}{CMT} & & \multicolumn{3}{c}{Metrics} \\
        \cline{6-8} 
         & & & & & Acc@both & Acc@base & Acc@novel  \\
         % \cline{7-8}
        \hline
        \multirow{2}{*}{\bluecheck} & \multirow{2}{*}{\bluecheck}  & \multirow{2}{*}{\bluecheck}  & \multirow{2}{*}{\bluecheck}  &$5-$shot & 60.54\% & 74.38\%  & 75.84\% \\
        & & & & $1-$shot & 54.97\% & 74.06\%  & 64.10\%\\
        \hline
        
        \multirow{2}{*}{\textcolor{red}{\xmark}} & \multirow{2}{*}{\bluecheck}  & \multirow{2}{*}{\bluecheck}  & \multirow{2}{*}{\bluecheck}  &$5-$shot &  58.92\% & 73.81\%  & 72.34\%\\
        & & & & $1-$shot & 53.35\% & 73.75\%  & 59.93\%\\
        \hdashline
        
        \multirow{2}{*}{\textcolor{red}{\xmark}} & \multirow{2}{*}{\textcolor{red}{\xmark}}  & \multirow{2}{*}{\bluecheck}  & \multirow{2}{*}{\bluecheck} &$5-$shot &  58.47\% & 73.96\%  & 71.67\%\\
        & & & & $1-$shot & 53.22\% & 73.67\%  & 59.46\%\\
        \hdashline
        
        \multirow{2}{*}{\textcolor{red}{\xmark}} & \multirow{2}{*}{\textcolor{red}{\xmark}}  & \multirow{2}{*}{\textcolor{red}{\xmark}}  & \multirow{2}{*}{\bluecheck}  &$5-$shot &   57.47\% & 70.96\%  & 69.67\%\\
        & & & & $1-$shot & 51.22\% & 71.67\%  & 57.46\%\\
        \hdashline
        
        \multirow{2}{*}{\textcolor{red}{\xmark}} & \multirow{2}{*}{\textcolor{red}{\xmark}}  & \multirow{2}{*}{\textcolor{red}{\xmark}}  & \multirow{2}{*}{\textcolor{red}{\xmark}}  &$5-$shot &  35.19\% & 62.98\%  & 40.52\% \\
        & & & & $1-$shot & 27.67\% & 61.72\%  & 32.83\%\\
        
        \hline
    \end{tabular}
    \label{tab:ablation}
    \vspace{-0.2cm}
\end{table}

\setlength{\tabcolsep}{6pt}
\begin{table}[t]
    \centering
    \caption{Performance with varying n-way/k-shot evaluation}
    \vspace{-0.05cm}
    \footnotesize
    \begin{tabular}{cccccc}
        \hline
         &   &   & \multicolumn{3}{c}{Metrics}\\
        \cline{4-6}
         & & & Acc@both & Acc@base & Acc@novel \\
        \hline
        \multirow{5}{*}{$5-$ way} 
         & \multicolumn{2}{c}{$1-$shot} & 54.97\% & 74.06\% & 64.10\% \\
        & \multicolumn{2}{c}{$5-$shot} & 60.54\% & 74.38\% & 75.84\%  \\
        & \multicolumn{2}{c}{$10-$shot} & 61.61\% & 74.14\%  & 76.95\%\\
         & \multicolumn{2}{c}{$15-$shot} & 62.08\% & 73.95\%  & 77.48\%\\
          & \multicolumn{2}{c}{$20-$shot} & 62.35\% & 74.83\%  & 78.35\%\\
        \hdashline
        \multirow{5}{*}{$10-$ way} 
         & \multicolumn{2}{c}{$1-$shot} & 43.62\% & 73.24\% & 47.31\% \\
        & \multicolumn{2}{c}{$5-$shot} & 51.82\% & 73.37\% & 59.97\%  \\
        & \multicolumn{2}{c}{$10-$shot} & 53.75\% & 73.54\%  & 61.21\%\\
         & \multicolumn{2}{c}{$15-$shot} & 55.46\% & 73.38\% & 62.74\% \\
          & \multicolumn{2}{c}{$20-$shot} & 57.58\% & 73.23\% & 64.37\%  \\
        
      \hline
    \end{tabular}
    \label{tab:exemplar_results}
        \vspace{-0.5cm}
\end{table}

\vspace{0.05cm}
\noindent \textbf{Effect of the number of sketch-exemplars:} Next, to investigate how the number of classes and samples affect the overall model performance, we evaluate the framework by varying the number of shots from $\{1, 5, 10, 15, 20\}$ and the number of ways only from $\{5, 10\}$. We depict the corresponding results in Table \ref{tab:exemplar_results}. We infer that larger way hurts the performance because of the ambiguity created by the new classes, while the model's performance increases when training with more number of samples. This portrays the potency of our proposed framework for other CIL variants.
\vspace{0.05cm}

\noindent \textbf{Comparison with text as support-set:} \cut{Here, we inspect the results by using the sketches as class support in comparison to text. We make use of the popular word embedding models, such as Word2Vec \cite{mikolov2013efficient}, and GloVe \cite{pennington2014glove} to generate class label representations.} 
% In order to compare our approach with text-based support set, we make use of the popular word embedding models, such as Word2Vec \cite{mikolov2013efficient}, and GloVe \cite{pennington2014glove} to generate class label representations.
In order to compare our approach with text-based support set, we use the word embeddings from Word2Vec \cite{mikolov2013efficient} and GloVe \cite{pennington2014glove} to generate class representations.
We delineated the results in Table \ref{tab:text_results}. The fine-grained nature of sketches helped surpass text results by a wide margin, showing its efficacy as the class support and a possible substitute to photos. \cut{The sketches' fine-grained nature enabled them outperform text results by a large margin, demonstrating its usefulness as a class support and a viable photo alternative.}
\vspace{0.05cm}

\begin{figure}[t]
	\begin{center}
		\includegraphics[width=1\linewidth]{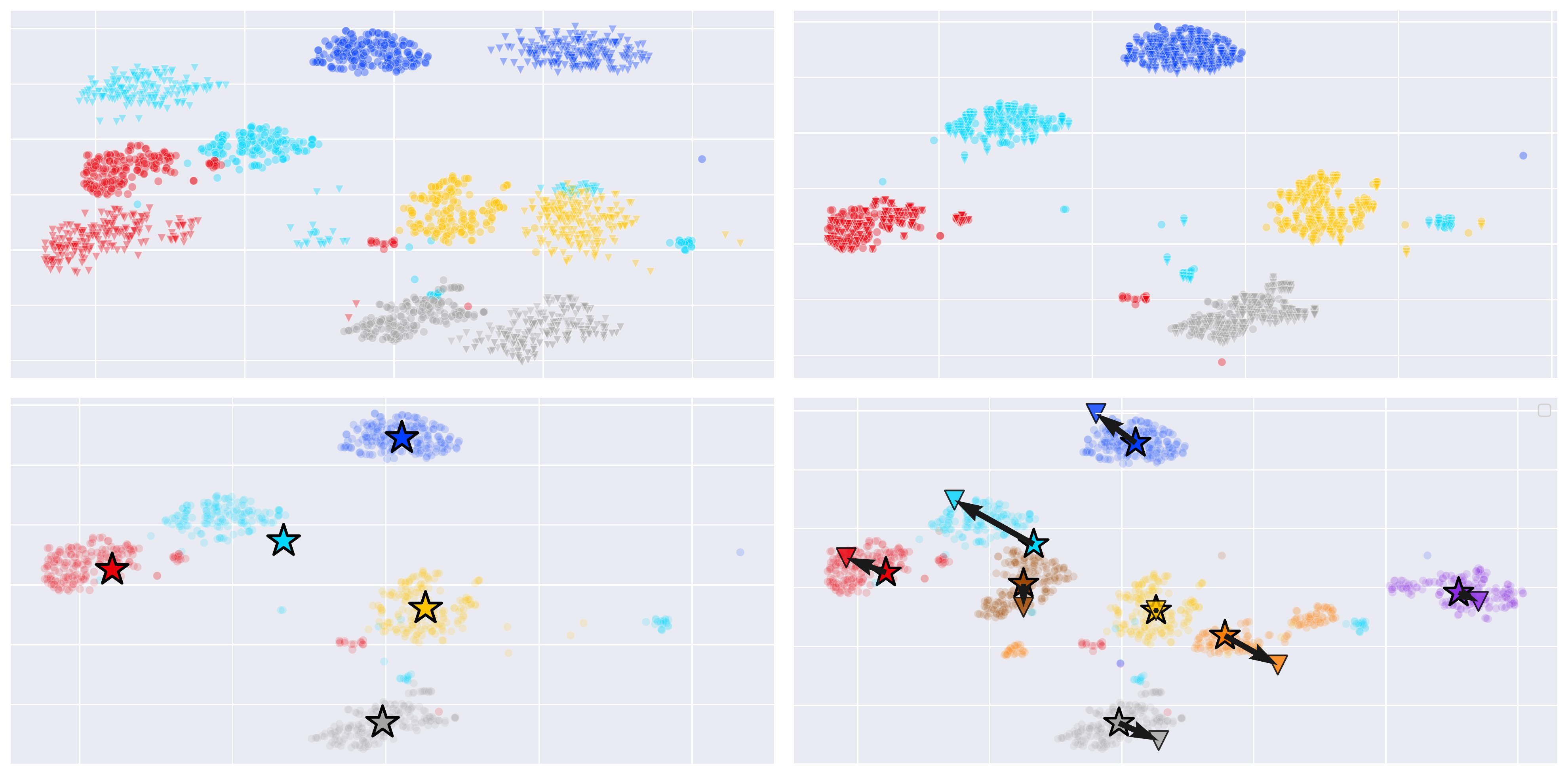} \\
		\vspace{-1.65in}
		\text{\hspace{-1.4in} \small{(a)} \hspace{1.4in} \small{(b)}}\\[0.65in]
		\text{\hspace{-1.4in} \small{(c)} \hspace{1.4in} \small{(d)}}\\
		\vspace{0.7in}
	\end{center}
	\vspace{-0.15in}
	\caption{t-SNE Plots: (a) Photo ($\circ$) and Sketch ($\nabla$) on the common embedding space of $\mathcal{F}_{\theta}$ using naive baseline. (b) Photo ($\circ$) and Sketch ($\nabla$) on the shared embedding space of $\mathcal{F}_{\theta}$ obtained \emph{using} our framework. This can be inferred that our method aligns the both modalities better by minimising the domain gap. 
(c) Base Classes (d) DIY-FSCIL. Here, deep-colour points are class prototypes, light-colour ones show the distribution of real data, star represents class representations during 1st stage, delta (bold) represents the refined vectors during incremental stage, and black arrow indicates the change in weights. Our DIY-FSCIL pushes the classifier weights away from the uncertain areas, resulting in better decision boundaries. Zoom in for better view.}

	\label{fig:tsne_plots}
	\vspace{-.4cm}
\end{figure}

\noindent \textbf{Visualisation of GAT refined features:}
With t-SNE \cite{van2008visualizing}, we visualise class representation weight vectors and classifier weights in a low-dimension space. We exhibit the results for the two configurations -- (i) with GAT, and (ii) without GAT. For this study, five classes are chosen randomly as the base classes, and five additional classes are added as incremental classes. As evident from Fig. \ref{fig:tsne_plots}, during incremental setup, the GAT module refines weights efficiently to push the classifier weights away from the uncertain areas, resulting in better decision boundary.

\setlength{\tabcolsep}{9pt}
\begin{table}[t]
    \centering
    \caption{Comparative study between \emph{sketch} vs \emph{text} for support set}
    \vspace{-0.05cm}
    \footnotesize
    \begin{tabular}{cccc}
        \hline
         &   \multicolumn{3}{c}{One-shot learning}\\
        \cline{2-4}
         & Acc@both & Acc@base & Acc@novel \\
        \hline
         Text (Word2Vec) & 22.85\% & 73.98\%  & 26.15\%\\
         Text (GloVe) & 22.80\% & 74.04\%  & 26.85\%\\
         Sketch (Ours) & 54.97\% & 74.06\%  & 64.10\%\\
       \hline
    \end{tabular}
    \label{tab:text_results}
        \vspace{-0.1cm}
\end{table}

\vspace{-0.3cm}
\section{Conclusion}
\vspace{-0.2cm}
% Incorporate doodling, privacy just by ob-serving a few sketchesdoodledby users themselves
In this paper, we have introduced a novel framework for few shot class incremental learning without violating the data privacy and ethical norms. This method also empowers the users to construct novel categories just by providing a few imaginative sketches doodled by themselves. The proposed framework unifies Knowledge Distillation, Gradient Consensus, and Graph Attention Networks to handle this newly proposed DIY-FSCIL paradigm. The effectiveness of the framework is validated by various experiments on the Sketchy dataset. Our framework is also extendable to other IL methods beyond the CIL used in this study. 

{\small
\bibliographystyle{ieee_fullname}
\bibliography{main}

\begin{thebibliography}{10}\itemsep=-1pt

\bibitem{aljundi2018memory}
Rahaf Aljundi, Francesca Babiloni, Mohamed Elhoseiny, Marcus Rohrbach, and
  Tinne Tuytelaars.
\newblock Memory aware synapses: Learning what (not) to forget.
\newblock In {\em ECCV}, 2018.

\bibitem{bhunia2021more}
Ayan~Kumar Bhunia, Pinaki~Nath Chowdhury, Aneeshan Sain, Yongxin Yang, Tao
  Xiang, and Yi-Zhe Song.
\newblock More photos are all you need: Semi-supervised learning for
  fine-grained sketch based image retrieval.
\newblock In {\em CVPR}, 2021.

\bibitem{bhunia2021vectorization}
Ayan~Kumar Bhunia, Pinaki~Nath Chowdhury, Yongxin Yang, Timothy~M Hospedales,
  Tao Xiang, and Yi-Zhe Song.
\newblock Vectorization and rasterization: Self-supervised learning for sketch
  and handwriting.
\newblock In {\em CVPR}, 2021.

\bibitem{sketchxpixelor}
Ayan~Kumar Bhunia, Ayan Das, Umar~Riaz Muhammad, Yongxin Yang, Timothy~M.
  Hospedales, Tao Xiang, Yulia Gryaditskaya, and Yi-Zhe Song.
\newblock Pixelor: A competitive sketching ai agent. so you think you can beat
  me?
\newblock In {\em ACM TOG}, 2020.

\bibitem{strokesubset}
Ayan~Kumar Bhunia, Subhadeep Koley, Abdullah Faiz Ur~Rahman Khilji, Aneeshan
  Sain, Pinaki~Nath Chowdhury, Tao Xiang, and Yi-Zhe Song.
\newblock Sketching without worrying: Noise-tolerant sketch-based image
  retrieval.
\newblock In {\em CVPR}, 2022.

\bibitem{bhunia2020sketch}
Ayan~Kumar Bhunia, Yongxin Yang, Timothy~M Hospedales, Tao Xiang, and Yi-Zhe
  Song.
\newblock Sketch less for more: On-the-fly fine-grained sketch-based image
  retrieval.
\newblock In {\em CVPR}, 2020.

\bibitem{chaudhry2018efficient}
Arslan Chaudhry, Marc'Aurelio Ranzato, Marcus Rohrbach, and Mohamed Elhoseiny.
\newblock Efficient lifelong learning with a-gem.
\newblock In {\em ICLR}, 2018.

\bibitem{chen2019closer}
Wei-Yu Chen, Yen-Cheng Liu, Zsolt Kira, Yu-Chiang~Frank Wang, and Jia-Bin
  Huang.
\newblock A closer look at few-shot classification.
\newblock In {\em ICLR}, 2019.

\bibitem{cheraghian2021semantic}
Ali Cheraghian, Shafin Rahman, Pengfei Fang, Soumava~Kumar Roy, Lars Petersson,
  and Mehrtash Harandi.
\newblock Semantic-aware knowledge distillation for few-shot class-incremental
  learning.
\newblock In {\em CVPR}, 2021.

\bibitem{PartialSBIR}
Pinaki~Nath Chowdhury, Ayan~Kumar Bhunia, Viswanatha~Reddy Gajjala, Aneeshan
  Sain, Tao Xiang, and Yi-Zhe Song.
\newblock Partially does it: Towards scene-level fg-sbir with partial input.
\newblock In {\em CVPR}, 2022.

\bibitem{collomosse2019livesketch}
John Collomosse, Tu Bui, and Hailin Jin.
\newblock Livesketch: Query perturbations for guided sketch-based visual
  search.
\newblock In {\em CVPR}, 2019.

\bibitem{dey2019doodle}
Sounak Dey, Pau Riba, Anjan Dutta, Josep Llados, and Yi-Zhe Song.
\newblock Doodle to search: Practical zero-shot sketch-based image retrieval.
\newblock In {\em CVPR}, 2019.

\bibitem{dong2021few}
Songlin Dong, Xiaopeng Hong, Xiaoyu Tao, Xinyuan Chang, Xing Wei, and Yihong
  Gong.
\newblock Few-shot class-incremental learning via relation knowledge
  distillation.
\newblock In {\em AAAI}, 2021.

\bibitem{dutta2019semantically}
Anjan Dutta and Zeynep Akata.
\newblock Semantically tied paired cycle consistency for zero-shot sketch-based
  image retrieval.
\newblock In {\em CVPR}, 2019.

\bibitem{ganin2015unsupervised}
Yaroslav Ganin and Victor Lempitsky.
\newblock Unsupervised domain adaptation by backpropagation.
\newblock In {\em ICML}, 2015.

\bibitem{ge2020creative}
Songwei Ge, Vedanuj Goswami, C~Lawrence Zitnick, and Devi Parikh.
\newblock Creative sketch generation.
\newblock In {\em ICLR}, 2021.

\bibitem{gidaris2018dynamic}
Spyros Gidaris and Nikos Komodakis.
\newblock Dynamic few-shot visual learning without forgetting.
\newblock In {\em CVPR}, 2018.

\bibitem{he2015delving}
Kaiming He, Xiangyu Zhang, Shaoqing Ren, and Jian Sun.
\newblock Delving deep into rectifiers: Surpassing human-level performance on
  imagenet classification.
\newblock In {\em ICCV}, 2015.

\bibitem{hertzmann2020line}
Aaron Hertzmann.
\newblock Why do line drawings work? a realism hypothesis.
\newblock {\em Perception}, 2020.

\bibitem{hinton2015distilling}
Geoffrey Hinton, Oriol Vinyals, and Jeff Dean.
\newblock Distilling the knowledge in a neural network.
\newblock In {\em NeurIPS Deep Learning Workshop}, 2014.

\bibitem{hsu2018re}
Yen-Chang Hsu, Yen-Cheng Liu, Anita Ramasamy, and Zsolt Kira.
\newblock Re-evaluating continual learning scenarios: A categorization and case
  for strong baselines.
\newblock In {\em NeurIPS Continual Learning Workshop}, 2018.

\bibitem{hu2020sketch}
Conghui Hu, Da Li, Yongxin Yang, Timothy~M Hospedales, and Yi-Zhe Song.
\newblock Sketch-a-segmenter: Sketch-based photo segmenter generation.
\newblock {\em IEEE TIP}, 2020.

\bibitem{kirkpatrick2017overcoming}
James Kirkpatrick, Razvan Pascanu, Neil Rabinowitz, Joel Veness, Guillaume
  Desjardins, Andrei~A Rusu, Kieran Milan, John Quan, Tiago Ramalho, Agnieszka
  Grabska-Barwinska, et~al.
\newblock Overcoming catastrophic forgetting in neural networks.
\newblock {\em Proc. of the NAS}, 2017.

\bibitem{koch2015siamese}
Gregory Koch, Richard Zemel, Ruslan Salakhutdinov, et~al.
\newblock Siamese neural networks for one-shot image recognition.
\newblock In {\em ICML}, 2015.

\bibitem{kuzborskij2013n}
Ilja Kuzborskij, Francesco Orabona, and Barbara Caputo.
\newblock From n to n+ 1: Multiclass transfer incremental learning.
\newblock In {\em CVPR}, 2013.

\bibitem{li2018learning}
Da Li, Yongxin Yang, Yi-Zhe Song, and Timothy~M Hospedales.
\newblock Learning to generalize: Meta-learning for domain generalization.
\newblock In {\em AAAI}, 2018.

\bibitem{li2019episodic}
Da Li, Jianshu Zhang, Yongxin Yang, Cong Liu, Yi-Zhe Song, and Timothy~M
  Hospedales.
\newblock Episodic training for domain generalization.
\newblock In {\em ICCV}, 2019.

\bibitem{li2021deep}
Xiaoyu Li, Bo Zhang, Jing Liao, and Pedro Sander.
\newblock Deep sketch-guided cartoon video inbetweening.
\newblock {\em IEEE TVCG}, 2021.

\bibitem{li2017learning}
Zhizhong Li and Derek Hoiem.
\newblock Learning without forgetting.
\newblock {\em IEEE TPAMI}, 2017.

\bibitem{liu2017deep}
Li Liu, Fumin Shen, Yuming Shen, Xianglong Liu, and Ling Shao.
\newblock Deep sketch hashing: Fast free-hand sketch-based image retrieval.
\newblock In {\em CVPR}, 2017.

\bibitem{luo2020towards}
Ling Luo, Yulia Gryaditskaya, Yongxin Yang, Tao Xiang, and Yi-Zhe Song.
\newblock Towards 3d vr-sketch to 3d shape retrieval.
\newblock In {\em 3DV}, 2020.

\bibitem{mansilla2021domain}
Lucas Mansilla, Rodrigo Echeveste, Diego~H Milone, and Enzo Ferrante.
\newblock Domain generalization via gradient surgery.
\newblock In {\em ICCV}, 2021.

\bibitem{mikolov2013efficient}
Tomas Mikolov, Kai Chen, Greg Corrado, and Jeffrey Dean.
\newblock Efficient estimation of word representations in vector space.
\newblock {\em arXiv preprint arXiv:1301.3781}, 2013.

\bibitem{pang2019generalising}
Kaiyue Pang, Ke Li, Yongxin Yang, Honggang Zhang, Timothy~M Hospedales, Tao
  Xiang, and Yi-Zhe Song.
\newblock Generalising fine-grained sketch-based image retrieval.
\newblock In {\em CVPR}, 2019.

\bibitem{pang2017cross}
Kaiyue Pang, Yi-Zhe Song, Tony Xiang, and Timothy~M Hospedales.
\newblock Cross-domain generative learning for fine-grained sketch-based image
  retrieval.
\newblock In {\em BMVC}, 2017.

\bibitem{paszke2017automatic}
Adam Paszke, Sam Gross, Soumith Chintala, Gregory Chanan, Edward Yang, Zachary
  DeVito, Zeming Lin, Alban Desmaison, Luca Antiga, and Adam Lerer.
\newblock Automatic differentiation in {PyTorch}.
\newblock In {\em NeurIPS Autodiff Workshop}, 2017.

\bibitem{pennington2014glove}
Jeffrey Pennington, Richard Socher, and Christopher~D Manning.
\newblock Glove: Global vectors for word representation.
\newblock In {\em EMNLP}, 2014.

\bibitem{polikar2001learn++}
Robi Polikar, Lalita Upda, Satish~S Upda, and Vasant Honavar.
\newblock Learn++: An incremental learning algorithm for supervised neural
  networks.
\newblock {\em IEEE TSMC:C}, 2001.

\bibitem{icart2017}
Sylvestre-Alvise Rebuffi, Alexander Kolesnikov, Georg Sperl, and Christoph~H.
  Lampert.
\newblock icarl: Incremental classifier and representation learning.
\newblock In {\em CVPR}, 2017.

\bibitem{rezende2016one}
Danilo Rezende, Ivo Danihelka, Karol Gregor, Daan Wierstra, et~al.
\newblock One-shot generalization in deep generative models.
\newblock In {\em ICML}, 2016.

\bibitem{ribeiro2020sketchformer}
Leo Sampaio~Ferraz Ribeiro, Tu Bui, John Collomosse, and Moacir Ponti.
\newblock Sketchformer: Transformer-based representation for sketched
  structure.
\newblock In {\em CVPR}, 2020.

\bibitem{rosenfeld2018incremental}
Amir Rosenfeld and John~K Tsotsos.
\newblock Incremental learning through deep adaptation.
\newblock {\em IEEE TPAMI}, 2018.

\bibitem{rusu2018meta}
Andrei~A Rusu, Dushyant Rao, Jakub Sygnowski, Oriol Vinyals, Razvan Pascanu,
  Simon Osindero, and Raia Hadsell.
\newblock Meta-learning with latent embedding optimization.
\newblock In {\em ICLR}, 2019.

\bibitem{Sketch3T}
Aneeshan Sain, Ayan~Kumar Bhunia, Vaishnav Potlapalli, Pinaki~Nath Chowdhury,
  Tao Xiang, and Yi-Zhe Song.
\newblock Sketch3t: Test-time training for zero-shot sbir.
\newblock In {\em CVPR}, 2022.

\bibitem{sain2020cross}
Aneeshan Sain, Ayan~Kumar Bhunia, Yongxin Yang, Tao Xiang, and Yi-Zhe Song.
\newblock Cross-modal hierarchical modelling for fine-grained sketch based
  image retrieval.
\newblock In {\em BMVC}, 2020.

\bibitem{sain2021stylemeup}
Aneeshan Sain, Ayan~Kumar Bhunia, Yongxin Yang, Tao Xiang, and Yi-Zhe Song.
\newblock Stylemeup: Towards style-agnostic sketch-based image retrieval.
\newblock In {\em CVPR}, 2021.

\bibitem{sangkloy2016sketchy}
Patsorn Sangkloy, Nathan Burnell, Cusuh Ham, and James Hays.
\newblock The sketchy database: learning to retrieve badly drawn bunnies.
\newblock {\em ACM TOG}, 2016.

\bibitem{santoro2016meta}
Adam Santoro, Sergey Bartunov, Matthew Botvinick, Daan Wierstra, and Timothy
  Lillicrap.
\newblock Meta-learning with memory-augmented neural networks.
\newblock In {\em ICML}, 2016.

\bibitem{shen2018zero}
Yuming Shen, Li Liu, Fumin Shen, and Ling Shao.
\newblock Zero-shot sketch-image hashing.
\newblock In {\em CVPR}, 2018.

\bibitem{snell2017prototypical}
Jake Snell, Kevin Swersky, and Richard~S Zemel.
\newblock Prototypical networks for few-shot learning.
\newblock In {\em NeurIPS}, 2017.

\bibitem{song2018learning}
Jifei Song, Kaiyue Pang, Yi-Zhe Song, Tao Xiang, and Timothy~M Hospedales.
\newblock Learning to sketch with shortcut cycle consistency.
\newblock In {\em CVPR}, 2018.

\bibitem{song2017fine}
Jifei Song, Yi-Zhe Song, Tony Xiang, and Timothy~M Hospedales.
\newblock Fine-grained image retrieval: the text/sketch input dilemma.
\newblock In {\em BMVC}, 2017.

\bibitem{song2017deep}
Jifei Song, Qian Yu, Yi-Zhe Song, Tao Xiang, and Timothy~M Hospedales.
\newblock Deep spatial-semantic attention for fine-grained sketch-based image
  retrieval.
\newblock In {\em CVPR}, 2017.

\bibitem{tao2020few}
Xiaoyu Tao, Xiaopeng Hong, Xinyuan Chang, Songlin Dong, Xing Wei, and Yihong
  Gong.
\newblock Few-shot class-incremental learning.
\newblock In {\em CVPR}, 2020.

\bibitem{tripathi2020sketch}
Aditay Tripathi, Rajath~R Dani, Anand Mishra, and Anirban Chakraborty.
\newblock Sketch-guided object localization in natural images.
\newblock In {\em ECCV}, 2020.

\bibitem{van2008visualizing}
Laurens Van~der Maaten and Geoffrey Hinton.
\newblock Visualizing data using t-sne.
\newblock {\em JMLR}, 2008.

\bibitem{velivckovic2017graph}
Petar Veli{\v{c}}kovi{\'c}, Guillem Cucurull, Arantxa Casanova, Adriana Romero,
  Pietro Lio, and Yoshua Bengio.
\newblock Graph attention networks.
\newblock In {\em ICLR}, 2018.

\bibitem{vinyals2016matching}
Oriol Vinyals, Charles Blundell, Timothy Lillicrap, Daan Wierstra, et~al.
\newblock Matching networks for one shot learning.
\newblock In {\em NeurIPS}, 2016.

\bibitem{vuorio2019multimodal}
Risto Vuorio, Shao-Hua Sun, Hexiang Hu, and Joseph~J Lim.
\newblock Multimodal model-agnostic meta-learning via task-aware modulation.
\newblock In {\em NeurIPS}, 2019.

\bibitem{wang2021sketchembednet}
Alexander Wang, Mengye Ren, and Richard Zemel.
\newblock Sketchembednet: Learning novel concepts by imitating drawings.
\newblock In {\em ICML}, 2021.

\bibitem{wang2020generalizing}
Yaqing Wang, Quanming Yao, James~T Kwok, and Lionel~M Ni.
\newblock Generalizing from a few examples: A survey on few-shot learning.
\newblock {\em ACM CSUR}, 2020.

\bibitem{xie2021exploiting}
Minshan Xie, Menghan Xia, and Tien-Tsin Wong.
\newblock Exploiting aliasing for manga restoration.
\newblock In {\em CVPR}, 2021.

\bibitem{xing2015autocomplete}
Jun Xing, Li-Yi Wei, Takaaki Shiratori, and Koji Yatani.
\newblock Autocomplete hand-drawn animations.
\newblock {\em ACM TOG}, 2015.

\bibitem{xu2022deep}
Peng Xu, Timothy~M Hospedales, Qiyue Yin, Yi-Zhe Song, Tao Xiang, and Liang
  Wang.
\newblock Deep learning for free-hand sketch: A survey.
\newblock {\em TPAMI}, 2022.

\bibitem{xu2018sketchmate}
Peng Xu, Yongye Huang, Tongtong Yuan, Kaiyue Pang, Yi-Zhe Song, Tao Xiang,
  Timothy~M. Hospedales, Zhanyu Ma, and Jun Guo.
\newblock Sketchmate: Deep hashing for million-scale human sketch retrieval.
\newblock In {\em CVPR}, 2018.

\bibitem{yang2020deep}
Shuai Yang, Zhangyang Wang, Jiaying Liu, and Zongming Guo.
\newblock Deep plastic surgery: Robust and controllable image editing with
  human-drawn sketches.
\newblock In {\em ECCV}, 2020.

\bibitem{yelamarthi2018zero}
Sasi~Kiran Yelamarthi, Shiva~Krishna Reddy, Ashish Mishra, and Anurag Mittal.
\newblock A zero-shot framework for sketch based image retrieval.
\newblock In {\em ECCV}, 2018.

\bibitem{yu2016sketch}
Qian Yu, Feng Liu, Yi-Zhe Song, Tao Xiang, Timothy~M Hospedales, and
  Chen-Change Loy.
\newblock Sketch me that shoe.
\newblock In {\em CVPR}, 2016.

\bibitem{yu2020gradient}
Tianhe Yu, Saurabh Kumar, Abhishek Gupta, Sergey Levine, Karol Hausman, and
  Chelsea Finn.
\newblock Gradient surgery for multi-task learning.
\newblock In {\em NeurIPS}, 2020.

\bibitem{zhang2021few}
Chi Zhang, Nan Song, Guosheng Lin, Yun Zheng, Pan Pan, and Yinghui Xu.
\newblock Few-shot incremental learning with continually evolved classifiers.
\newblock In {\em CVPR}, 2021.

\bibitem{zhang2021sketch2model}
Song-Hai Zhang, Yuan-Chen Guo, and Qing-Wen Gu.
\newblock Sketch2model: View-aware 3d modeling from single free-hand sketches.
\newblock In {\em CVPR}, 2021.

\end{thebibliography}
}

\cleardoublepage

\onecolumn{}
\title{\vspace{-0.5cm}\Large{\textbf{Supplementary material for \\ Doodle It Yourself: Class Incremental Learning \\ by Drawing a Few Sketches}\vspace{-0.5cm}}}

% Authors at the same institution
%%%%%%%%% TITLE
 
% Authors at the same institution
% Authors at the same institution
\author{Ayan Kumar Bhunia\textsuperscript{1} \hspace{.3cm}  Viswanatha Reddy Gajjala$^{*}$  \hspace{.3cm}  Subhadeep Koley\textsuperscript{1,2}  \hspace{.2cm} Rohit Kundu$^{*}$ \\   \hspace{.2cm} Aneeshan Sain\textsuperscript{1,2}  \hspace{.2cm}   
Tao Xiang\textsuperscript{1,2}\hspace{.2cm}  Yi-Zhe Song\textsuperscript{1,2} \\
\textsuperscript{1}SketchX, CVSSP, University of Surrey, United Kingdom.  \\
\textsuperscript{2}iFlyTek-Surrey Joint Research Centre on Artificial Intelligence.\\
{\tt\small \{a.bhunia, s.koley, a.sain, t.xiang, y.song\}@surrey.ac.uk; viswanathareddy998@gmail.com} 
% \vspace{-1.0cm}
}

\maketitle

\section*{Evaluating with poor-quality sketches} 
For that, we further evaluated our Sketchy-trained model on 20 Quick-Draw (poor quality) classes without any further re-training. We obtain reasonable results without a major drop in accuracy (Table \ref{tab:maintable_sup}). We also separated the `hard' sketches in the Sketchy dataset as those having high entropy \cite{sangkloy2016sketchy}, and separately experimented on those. Results show Acc@both/Acc@base/Acc@novel (5-shot) of $55.24\%/ 69.43\%/ 71.06\%$ respectively, again not a significant drop compared with our final results. 

\setlength{\tabcolsep}{6pt}
\begin{table}[!hbt]
    \centering
    
\caption{Experiment results on QuickDraw}    
    
    \vspace{0.05cm}
    \footnotesize
    \begin{tabular}{cccccccccc}
        \hline
        %  &   \multicolumn{5}{c}{5-Shot Learning} \\
        % \cline{4-6}  \cline{7-11}
         & & Acc@both & Acc@base &Acc@novel \\
        \hline
        % 60.54\% & 74.38\% & 75.84
         \multicolumn{2}{c}{5-Shot Learning} & 55.24\% & 69.43\% & 71.06\% \\ % & 54.97 \% & 74.06\% & 64.10\%\\
        
        \multicolumn{2}{c}{1-Shot Learning} & 50.16\% & 69.12\% & 59.76\%\\
        
      \hline
    \end{tabular}
    \label{tab:maintable_sup}
        \vspace{-0.3cm}
\end{table}

\section*{Contribution of Gradient-Consensus} 

 While Gradient Consensus (GC) is used for both first and second stages of training, we report the results only with respect to stage-2 in Table 2. In particular, GC improves the base class accuracy of the stage-1 model by $2.72 \% $. We agree that the contribution of GC is minimal for stage-2, but removing GC from stage-1 would reduce all reported numbers of stage-2 due to conflicting gradient incurred by the large domain gap between sketch and photo.  Precisely, our final results without the stage-1 gradient consensus will reduce  Acc@both/Acc@base/Acc@novel (5-shot) to $57.21\%$/$70.13\%$/$71.12\%$ respectively (by a factor of $3\%-4\%$ decrements). This shows the impact of the GC on the overall framework. 

\section*{Limitations and future work}
 
{In summary, our limitations are (i) Acc@both and Acc@novel lag behind the upper bound by $9-10\%$, signifying scope for further improvement (ii) lacking explicit training protocol for better cross-style generalisation across sketches. 
}

\end{document}